\documentclass[journal]{IEEEtran}
\usepackage{graphics,graphicx}
\usepackage{amsmath,amssymb,amsthm}
\usepackage{multirow}
\usepackage{cases}
\usepackage{array}
\usepackage{algorithm,algorithmic}
\usepackage{color,colortbl,xcolor}
\usepackage[graphicx]{realboxes}
\usepackage{booktabs,url}
\usepackage[tight,footnotesize,sf,SF]{subfigure}
\usepackage{bigstrut}

\newcommand{\fref}[1]{Fig. \ref{#1}}
\newcommand{\sref}[1]{Section \ref{#1}}
\newcommand{\tref}[1]{TABLE \ref{#1}}
\newcommand{\eref}[1]{Eq. (\ref{#1})}

\begin{document}
\title{Coevolutionary Framework for Generalized Multimodal Multi-objective Optimization}

\author{Wenhua Li,
Xingyi Yao,
Kaiwen Li,
Rui Wang \emph{Senior Member}, 
Tao Zhang,
Ling Wang
\thanks{This work was supported by the National Science Fund for Outstanding Young Scholars (62122093), the National Natural Science Foundation of China (72071205), the Scientific key Research Project of National University of Defense Technology (ZZKY-ZX-11-04) 
and the Ji-Hua Laboratory Scientific Project (X210101UZ210).}
\thanks{Wenhua Li and Xingyi Yao are with the College of Systems Engineering, National University of Defense Technology, Changsha, China, 410073, e-mail: (liwenhua@nudt.edu.cn).}% <-this % stops a space
\thanks{Rui Wang and Tao Zhang (contribute equally to this work, Corresponding authors, email: ruiwangnudt@gmail.com) are with College of Systems Engineering, National University of Defense Technology, Changsha, China and Hunan Key Laboratory of Multi-energy System Intelligent Interconnection Technology, Changsha, 410073, China}% <-this % stops a space
\thanks{Ling Wang is with the Department of Automation, Tsinghua University, Beijing, 100084, P.R. China. }
}

\maketitle

\begin{abstract}
Most multimodal multi-objective evolutionary algorithms (MMEAs) aim to find all global Pareto optimal sets (PSs) for a multimodal multi-objective optimization problem (MMOP). However, in real-world problems, decision makers (DMs) may be also interested in local PSs. Also, searching for both global and local PSs is more general in view of dealing with MMOPs, which can be seen as a generalized MMOP. In addition, current state-of-the-art MMEAs exhibit poor convergence on high-dimension MMOPs. To address the above two issues, we present a novel coevolutionary framework (CoMMEA) to better produce both global and local PSs, and simultaneously, to improve the convergence performance in dealing with high-dimension MMOPs. Specifically, the CoMMEA introduces two archives to the search process, and coevolves them simultaneously through effective knowledge transfer. The convergence archive assists the CoMMEA to quickly approaching the Pareto optimal front (PF). The knowledge of the converged solutions is then transferred to the diversity archive which utilizes the local convergence indicator and the $\epsilon$-dominance-based method to obtain global and local PSs effectively. Experimental results show that CoMMEA is competitive compared to seven state-of-the-art MMEAs on fifty-four complex MMOPs.
\end{abstract}

\IEEEpeerreviewmaketitle

\section{Introduction}
\label{sec_introduction}

In the real world, most engineering optimization problems need to minimize more than one objective function simultaneously, which is often termed multi-objective optimization problems (MOPs). Mathematically, an MOP can be defined as follows:

\begin{equation}
\begin{gathered}
\text{Minimize}~~F(\mathbf{x}) = \{f_1(\mathbf{x}),f_2(\mathbf{x}),\cdots, 
f_m(\mathbf{x}) \}, \\
s.t.~~~~ \mathbf{x} = (x_1,x_2,\ldots, x_n) \in \Omega,
\end{gathered}
\label{eq_MOP}
\end{equation}
where $\Omega$ represents the decision space, $m$ represents the number of objectives, and $\mathbf{x}$ is a decision vector made up of $n$ decision variables $x_i$. A solution, $\mathbf{x_a}$, is considered to Pareto dominate another solution, $\mathbf{x_b}$, iff $\forall i=1,2,...,m, f_i(\mathbf{x_a}) \leq f_i(\mathbf{x_b})$ and $\exists j=1,2,...,m, f_j(\mathbf{x_a}) < f_j(\mathbf{x_b})$. Furthermore, a Pareto optimal solution is a solution that is not Pareto dominated by any other solution. The set of Pareto optimal solutions is called a Pareto optimal set (PS). The image of the PS in objective space is known as the Pareto optimal front (PF).

To effectively solve MOPs, a number of multi-objective evolutionary optimization algorithms (MOEAs) have been developed which aim to obtain a set of well-distributed solutions that are also close to the PF. In general, for most of MOEAs, maintaining diversity in the objective space is adopted as the second criterion (while convergence is taken as the first criterion) to select the offspring solutions. Therefore, the state-of-the-art MOEAs, e.g., NSGA-II \cite{deb2002fast}, MOEA/D\cite{zhang2007moea}, PICEAs\cite{wang2012preference}, can obtain solutions with good convergence and diversity for MOP benchmarks.

\begin{figure}[tbh]
\begin{center}
\includegraphics[width=3.5in]{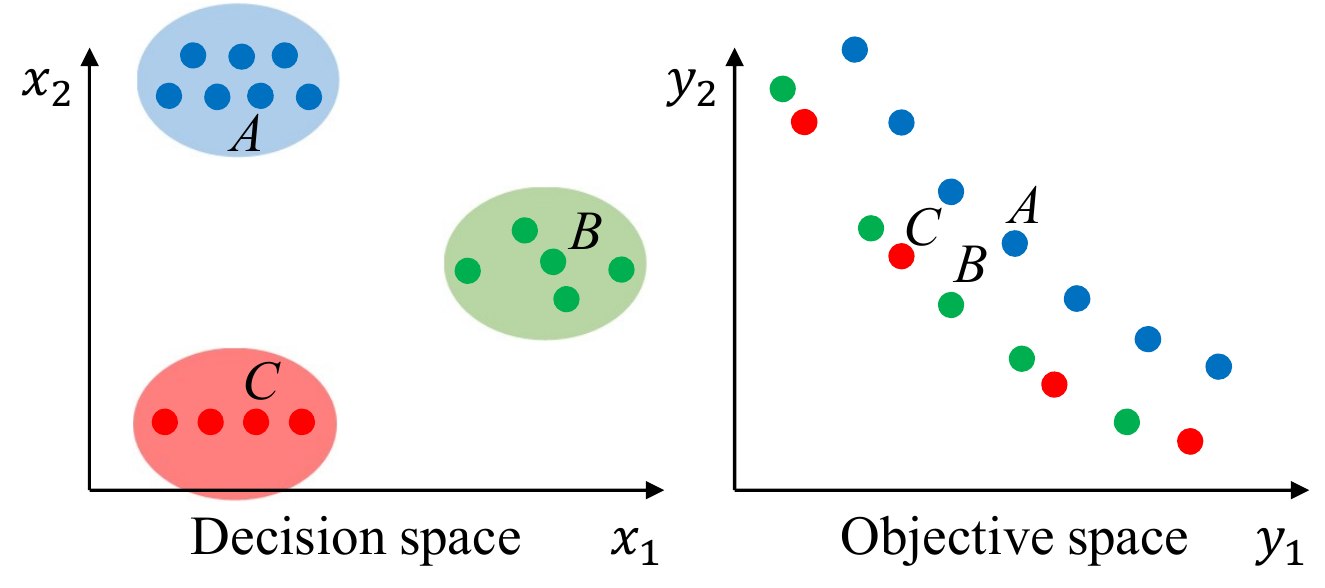}
\caption{Illustration of the diagram of the coevolutionary framework for generalized MMOPs (GMMOPs).}
\label{fig_gmmop}
\end{center}
\end{figure}

However, multimodal multi-objective optimization problems (MMOPs) have posed great challenges to the state-of-the-art MOEAs. MMOPs are problems with several distinct optimal solutions that have very similar objective values, which are frequently encountered in the fields of protein design \cite{lapizco2010particle}, drug structures optimization \cite{sakhovskiy2022multimodal}, path planning \cite{yao2022multimodal}, feature selection \cite{yue2019multimodal} and so on. For these problems, obtaining as many diversified optimal solutions as possible is essential. Take the drug design problem as an example, for two different drug structures that have similar objective values, if one structure is hard to construct while the other one can be easily obtained, then the decision makers (DMs) would be more likely to choose the latter even if its objective values are a bit inferior to the former. \fref{fig_gmmop} illustrates the generalized MMOPs (GMMOPs), where solutions $A$, $B$ and $C$ are far from each other in the decision space but are close to each other in the objective space. In this situation, solutions $B$ and $C$ correspond to global PSs while solution $A$ is a local PS.

MMOPs have been a hot topic in the evolutionary computation community. Typically, MMOPs refer to problems that have more than one global/local PS \cite{tanabe2019review}. The main tasks for most MMEAs are to obtain all of the equivalent global PSs and to maintain the uniform distribution of solutions in the decision space \cite{grimme2021peeking, javadi2021using, dang2022dynamic, li2021two}. That is, they try to obtain the red and green regions in \fref{fig_gmmop} while discarding the blue region. However, as reported in \cite{li2022hierarchy, li2022local}, a more general case for real-world problems is that different PSs correspond to PFs that are NOT equal but are close to each other in the objective space, i.e., one PF is slightly worse than another. In this study, we regard obtaining both global and local PSs as the aim for GMMOPs, which has not been well studied yet. 

The limitations of existing MMEAs are as follows. First, the poor convergence of MMEAs has hindered their application to real-world problems. Second, in previous literature, MMOP test suites contained only two or three decision variables. The main reason is that multimodality needs to be intuitively observed in the decision space. However, such problems are insanely simple for general MOEAs, that is, the true PF can be easily obtained within a few generations. It is reported in \cite{li2022multimodal} that existing MMEAs show poor performance as the number of decision variables increases. Third, unless the DMs know the problems well, it is usually hard to determine whether a problem is an MMOP or not. As a result, general MOEAs would be the first choice rather than MMEAs. Finally, the need to obtain all global and local PSs contradicts the goal of general MOEAs, which is to obtain a set of diverse solutions with good convergence.

To address the issues, we propose a $\epsilon$-dominance \cite{loridan1984varepsilon} embeded coevolutionary framework based on two archives to handle GMMOPs. Specifically, a convergence archive is utilized to enhance the searching ability and to lead the algorithm to quickly converge to the true PF. A diversity archive updated by a local convergence indicator as well as the $\epsilon$-dominance is adopted to obtain both global and local PSs. It is worth mentioning that standard MOEAs can be readily implemented into this framework. The CoMMEA aims to obtain all global and local PSs while improving the convergence of MMEAs. In a nutshell, the main contributions are as follows:

\begin{itemize}
\item A coevolutionary framework for dealing with GMMOPs is proposed, which utilizes a convergence archive to effectively lead candidate solutions to approach the true PF, and a diversity archive to maintain the diversity of solutions in both the decision and objective spaces. The coevolutionary framework takes full advantage of these two archives. As a result, both global and local PSs with good convergence can be obtained effectively.
\item An adaptive local convergence indicator ($I_{LC}$) that is more robust in terms of forming stable niches is proposed. Therefore, CoMMEA is able to solve high-dimension MMOPs (MMOPs with many decision variables), obtaining both global and local PSs.
\item The performance of CoMMEA is verified on different sets of MMOPs (54 test problems in total), including easy MMOPs with 2-4 decision variables, imbalance MMOPs, high-dimension MMOPs and MMOPLs. Experimental results indicate that CoMMEA outperforms seven state-of-the-art MMEAs on most of the chosen test problems.
\end{itemize}

The rest of this study is structured as follows: \sref{sec_prework} reviews related works briefly in literature, followed by \sref{sec_method}, which illustrates the CoMMEA in detail. Then in \sref{sec_exp}, several state-of-the-art MMEAs and MMOP test suites are chosen to examine the performance of the CoMMEA. In addition, the analysis of parameter $\epsilon$ and the search behaviors of the two archives in CoMMEA are discussed in detail. Finally, \sref{sec_con} concludes the paper and identifies future directions.

\section{Preliminary work}
\label{sec_prework}
\subsection{Multimodal multi-objective optimization}

Multimodal multi-objective optimization (MMO) has attracted increasing attentions over the past two decades, which aims to find as many equivalent optimal solutions as possible. To accomplish this goal, several MMEAs have been proposed, e.g., Omni-optimizer \cite{deb2005omni,deb2008omni}, double-niched evolutionary algorithm (DNEA) \cite{liu2018double}, decision space-based niching NSGA-II (DN-NSGA-II) \cite{liang2016multimodal}, MO\_Ring\_PSO\_SCD \cite{yue2017multiobjective}, self-organizing quantum-inspired particle swarm optimization algorithm (MMO\_SO\_QPSO) \cite{li2021handling}, MMEA-WI \cite{li2021weighted} and so on. In addition, to better evaluate the performance, a number of MMOP benchmarks have been proposed. For example, the MMF test suite \cite{liang2016multimodal}, which contains two decision variables and two objectives, can be used to intuitively present the search behaviors of MMEAs. The IDMP test suite \cite{liu2019handling} has multiple PSs with various difficulties. IDMPs have posed challenges to early MMEAs, that is, the MMEAs can only find one single PS. Liang et al. \cite{liang2019problem} have made a lot of efforts in designing new benchmarks, and they have raised MMO competitions, i.e., the IEEE CEC MMO competition serials.

Recently, a remarkable review study has been made by Tanabe et al. \cite{tanabe2019review}, where the importance of finding both global and local PSs is pointed out. The requirement of having local PSs is that DMs may prefer solutions with slightly inferior objectives values but higher robustness. TO find solutions with acceptable convergence, $P_{Q,\epsilon}$-MOEA \cite{schutze2011computing} uses the $\epsilon$-dominance to find solutions that have acceptable convergence, while DNEA-L \cite{liu2019searching} introduces a multi-front archive update method. To better solve the IEEE CEC 2020 test suite, Lin et.al \cite{lin2020multimodal} proposed a dual clustering in the decision and objective spaces, named MMOEA/DC. In addition, a clearing-based evolutionary algorithm with layer-to-layer evolution strategy (CEA-LES) \cite{wang2021clearing} is proposed. A hierarchy ranking based evolutionary algorithm (HREA) \cite{li2022hierarchy} is proposed wherein a novel test suite (IDMP\_e) with both global and local PSs is proposed. Compared to MMEAs that focus only on obtaining global PSs, there are few studies that aim to find both global and local PSs. Overall, multimodal multi-objective optimization has been studied a lot, however, MMEAs still face challenges in dealing with large-scale MMOPs, and in particular, few works have investigated GMMOPs.

\subsection{$\epsilon$-dominance}
The $\epsilon$-dominance was first proposed by Laumanns et al. \cite{laumanns2002combining} in 2002, which aims to improve the solution diversity of MOEAs. It can be defined as follows.

\textbf{Definition 1 ($\epsilon$-dominance)} Let $\mathbf{x_1},\mathbf{x_2}\in \Omega$. Then $\mathbf{x_1}$ is said to $\epsilon$-dominate $\mathbf{x_2}$ for some $\epsilon>0$ (in short $\mathbf{x_1}\succ _\epsilon \mathbf{x_2}$), $iff$ for all $i\in \{1,...,m\}$
\begin{equation}
(1+\epsilon)\cdot f_i(\mathbf{{x}_1})\geq f_i(\mathbf{{x}_2})
\end{equation}

With the definition of $\epsilon$-dominance, the $\epsilon$-approximate Pareto set can be defined as follows.

\textbf{Definition 2 ($\epsilon$-approximate Pareto set)} Let $F\subseteq \mathbb{R}_+^m$ be a set of vectors and $\epsilon>0$. Then a set $F_{\epsilon}$ is called an $\epsilon$-approximate Pareto set of $F$, if any vector $g\in F$ is $\epsilon$-dominated by at least one vector $f\in F_{\epsilon}$, i.e.
\begin{equation}
\forall g\in F : \exists f\in F_{\epsilon} \quad \text{such that} \quad f\succ _\epsilon g
\end{equation}

\begin{figure}[tbh]
\begin{center}
\includegraphics[width=3in]{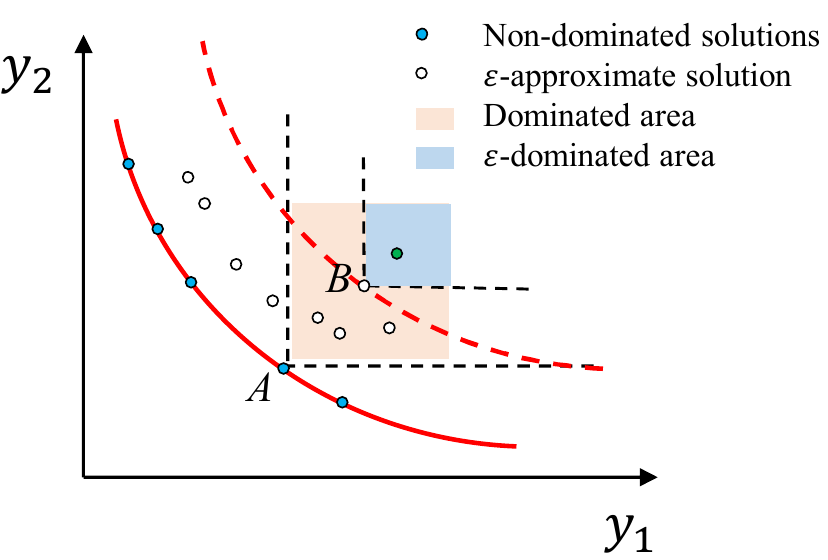}
\caption{Illustration of $\epsilon$-dominance and $\epsilon$-approximate Pareto set.}
\label{fig_epsdom}
\end{center}
\end{figure}

As shown in \fref{fig_epsdom}, we assume that the red line is the true PF while point $A$ is a non-dominated solution and for all $i\in \{1,...,m\}$,
\begin{equation}
(1+\epsilon)\cdot f_i(\mathbf{A})\geq f_i(\mathbf{B})
\end{equation}
then we can identify the dominated area by point $A$ (highlighted by orange background) and the $\epsilon$-dominated area (highlighted by blue background) by point $A$, respectively. Therefore, the solutions obtained through $\epsilon$-dominance is a strip area between the true PF and the $\epsilon$-dominated area. 

The use of $\epsilon$-dominance is discussed in detail in \cite{horoba2008benefits}, where the authors pointed out that it can significantly help to reduce the runtime of an MOEA till a good approximation of the PF is achieved. In addition, the diversity of the population can be well enhanced. However, when the number of feasible objective vectors is small, it will slow down the optimization process drastically. Therefore, how to utilize the $\epsilon$-dominance to enhance diversity and maintain fast convergence speed is an interesting topic.

\subsection{Local optimal solutions}
\begin{figure}[tbh]
	\begin{center}
		\includegraphics[width=2in]{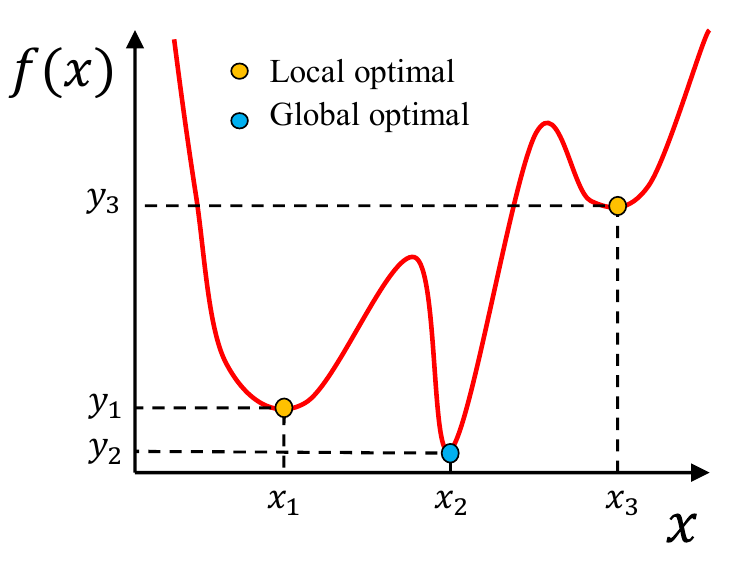}
		\caption{Illustration of global and local optima.}
		\label{fig_local}
	\end{center}
\end{figure}

In order to obtain both global and local PSs for GMMOPs, the local convergence strategy is a potential method. \fref{fig_local} illustrates global and local optima with only one decision variable and one objective, where we can see that there is one global optimum $x_2$ and two local optima $x_1$ and $x_3$.

As we can see, the landscape around the global optimum is very steep. That is, a small disturbance in the decision space would cause a huge change in the objective space. For local optimum $x_1$, its surrounding landscape is smooth, which can be seen as a robust solution. We assume that $x$ is the workpiece length and $f(x)$ is the performance evaluation of the workpiece. In this case, the need for mechanical processing accuracy for $x_1$ is low, making a more preferred solution possible, demonstrating the importance of achieving both global and local efficiencies. In addition, for solution $x_3$, the performance is poor, which should be discarded. To better illustrate the concept, we define the local non-dominated solution as follows:

\textbf{Definition 3 (Local non-dominated solution)} Let $\mathbf{x}\in \Omega$ and $\Omega_{\mathbf{x}}$ be the neighbor solution set of $\mathbf{x}$. Then $\mathbf{x}$ is said to be a local non-dominated solution iff $\mathbf{x}$ is a non-dominated solution inside $\Omega_{\mathbf{x}}$.

As we can see, the definition of a local non-dominated solution contains the concept of neighbor solutions. Then, several approaches were proposed to better find the neighbor solutions for a certain individual, e.g., niching strategy \cite{liu2019searching}, distance-based method \cite{li2022hierarchy}, clustering method \cite{lin2020multimodal} and zoning search \cite{fan2019solving, fan2021zoning}. For the distance-based method, a solution $\mathbf{y}$ is said to be a neighbor solution of $\mathbf{x}$ if the Euclidean distance in the decision space between these two solutions is less than a predefined threshold $R$. Distinctively, if $R=0$, then all solutions can be seen as local non-dominated solutions; if $R=inf$, then local non-dominated solutions are effectively global non-dominated. Since $R$ is defined by DMs, it is hard to find the most suitable way for all problems. The clustering method tries to divide solutions into several groups and consider solutions inside the same group as neighbor solutions. However, the accuracy of classification greatly affects the performance of algorithms. On the other hand, the zoning method simply divides decision space into several sub-regions to form niches. The performance of the zoning method decreases as the number of decision variables increases. Therefore, figuring out a robust method to find neighbor solutions is still a challenging issue so far.
%obtaining stable niches to calculate the local non-dominated solutions is an interesting and important issue for the MMO community so far[XX这句话有什么意义？？].

\section{Coevolutionary framework for GMMOPs}
\label{sec_method}
Obtaining as many well-distributed optimal solutions as possible is the main task for most MMEAs. To achieve this goal, existing MMEAs prefer embedding crowding distance in the decision space into the environmental selection strategy. Such approaches are effective in obtaining multiple equivalent PSs for the existing benchmarks, e.g., the MMF test suite and Omni test problems. However, the convergence in the objective space and diversity in the decision space are somehow in conflict. Therefore, overemphasizing the diversity in the decision space may lead to poor convergence for problems with many decision variables. To better balance the convergence and the diversity in the decision space, inspired by \cite{ming2021dual, tian2020coevolutionary, li2021sizing}, we propose a novel coevolutionary framework with two archives. Specifically, the convergence archive adopts the convergence-first strategy (standard MOEAs) to perform the environmental selection, aiming to find the true global PF/PS of the MMOPs. The diversity archive utilizes a local convergence indicator to maintain both global and local PSs, and adopts the $\epsilon$-dominance to control the convergence of the local PS. Therefore, the proposed framework can be applied to other MOEAs to enhance their convergence, obtaining as many global and local PSs as possible during the evolution.

\subsection{Framework}

\renewcommand{\algorithmicrequire}{\textbf{Input:}} % Use Input in the 
%format of Algorithm
\renewcommand{\algorithmicensure}{\textbf{Output:}} % Use Output in the 
%format of Algorithm
\begin{algorithm}[htb]
\caption{General Framework of CoMMEA}
\label{alg_framework}
\begin{algorithmic}[1]
\REQUIRE {Maximum generations $MaxGen$, population size $N$, parameter for $\epsilon$-approximate PS $\epsilon$}
\ENSURE {Diversity archive $DA$}
%\BlankLine
\STATE $CA \leftarrow Initialization(N)$ /* Generate the convergence archive. */
\STATE $DA \leftarrow Initialization(N)$ /* Generate the diversity archive. */
\STATE $FitnessC \leftarrow CalFitness(CA,true)$ /* Calculate the convergence fitnesses for parents selection. */
\STATE $FitnessD \leftarrow CalFitness(DA,false)$

\WHILE{$NotTerminated$}
\STATE $ParentC \leftarrow TournamentSel(CA,FitnessC,N/2)$
\STATE $ParentD \leftarrow TournamentSel(DA,FitnessD,N)$ /* Select $N$ parents through tournament selection method. */
\STATE $OffC \leftarrow Variation(ParentC)$ /* Generate offspring. */
\STATE $OffD \leftarrow Variation(ParentD)$

\STATE $[CA,FitnessC] \leftarrow EnvSel1(CA,OffC,$ $OffD,N)$ /* Use a normal MOEA to select a new population for convergence archive. */ 
\STATE $[DA,FitnessD] \leftarrow EnvSel2(DA,OffC,OffD,$ $N,\epsilon)$ /* Use $\epsilon$-dominance based method to select new population for diversity archive. */
\ENDWHILE
\end{algorithmic}
\end{algorithm}
\begin{figure}[tbh]
\begin{center}
\includegraphics[width=3.5in]{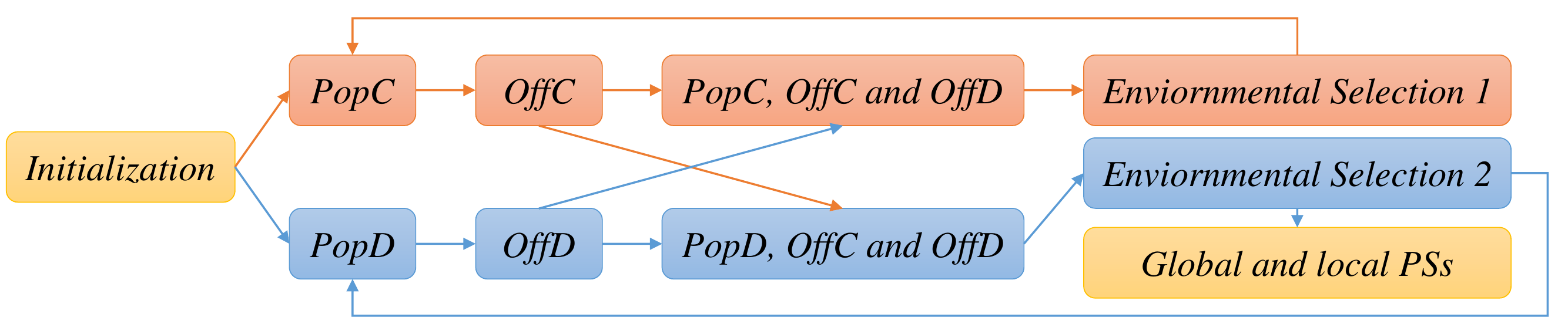}
\caption{Diagram of the coevolutionary framework for GMMOPs.}
\label{fig_framework}
\end{center}
\end{figure}

The framework of CoMMEA is described in Algorithm \ref{alg_framework} and \fref{fig_framework}. As we can see, two archives (convergence archive $CA$ and diversity archive $DA$) are utilized to better balance the convergence in the objective space and the diversity in the decision space. In the beginning, the same initialization method is utilized to generate $CA$ and $DA$, which are then used to calculate the fitness for the parent's selection. During the evolution, parents are chosen from $CA$ and $DA$ to generate offspring separately, see $lines\ 6-9$. Then, the coevolutionary mechanism is adopted, where the offspring from both $CA$ and $DA$ are merged to separately update these two archives. Specifically, the environmental selection strategies from normal MOEAs can be used to update $CA$, while a novel $\epsilon$-dominance-based method is proposed to form the new $DA$. In this way, solutions with good convergence quality in $CA$ can lead the algorithm. On the other hand, solutions with good distribution in $DA$ can further improve the solution diversity in the convergence archive. Overall, the diagram of the proposed coevolutionary framework for solving GMMOPs is presented in \fref{fig_framework}. The coevolution between $CA$ and $DA$ is presented by the offspring sharing. It's worth mentioning that, the updating method for $CA$ could be various, which is provided by the existing MOEAs.

\subsection{Environmental selection for convergence archive}

The convergence archive pushes the individuals to quickly approach the true PF, providing useful knowledge to the diversity archive so as to guide the search more effectively. Therefore, environmental selection strategies in current MOEAs can be utilized, e.g, the Pareto-dominance-based method, the indicator-based approach, and the decomposition-based method. In this work, we adopt the method proposed in SPEA2 \cite{zitzler2001spea2}.

\begin{algorithm}[htb]
\caption{Environmental selection for convergence archive}
\label{alg_env}
\begin{algorithmic}[1]
\REQUIRE {Convergence archive $CA$, offspring from convergence archive $OffC$, offspring from diversity archive $OffD$, population size $N$}
\ENSURE {Updated convergence archive $CA$, individual fitness $FitnessC$ }
\STATE $JointArc \leftarrow DA \cup OffC \cup OffD$
\STATE $FitnessC \leftarrow CalFitness(JointArc)$ /* Use fitness assignment method proposed in SPEA2 \cite{zitzler2001spea2}. */
\STATE $Next \leftarrow FitnessC < 1$ /* Obtain the current PF. */
\IF{$Num(Next)<N$}
\STATE $CrowdD \leftarrow CalCrowdDis(JointArc)$
\STATE $IDX \leftarrow sort(FitnessC,CrowdD)$
\STATE $CA \leftarrow(JointArc(IDX(1:N)))$
\ELSE
\STATE $CA \leftarrow Truncate(JointArc(Next),N)$
\ENDIF
\STATE $FitnessC \leftarrow CalFitness(CA)$
\end{algorithmic}
\end{algorithm}

In particular, the original population is combined with offspring to form a joint archive $JointArc$, which is then used to calculate the convergence fitness $FitnessC$ ($lines\ 1-2$). Notably, $FitnessC<1$ means that the individual is non-dominated. If the number of non-dominated solutions is less than the population size $N$, the first $N$ solutions sorted by $FitnessC$ and crowding distance are chosen to form the new population; if the value exceeds $N$, a crowding distance-based truncation method is used to keep the population size ($lines 4-9$).

\subsection{Environmental selection for diversity archive}
To find all global PSs and local PS with acceptable convergence, a parameter $\epsilon$ is introduced, which is equivalent to the definition of $\epsilon$-dominance. That is, if the problem is not a GMMOP, then CoMMEA will output the true PF and PSs; if the problem is a GMMOP, then all global PSs and all local PSs with $\epsilon$-acceptable quality will be obtained.

\begin{figure}[tbh]
\begin{center}
\includegraphics[width=3.5in]{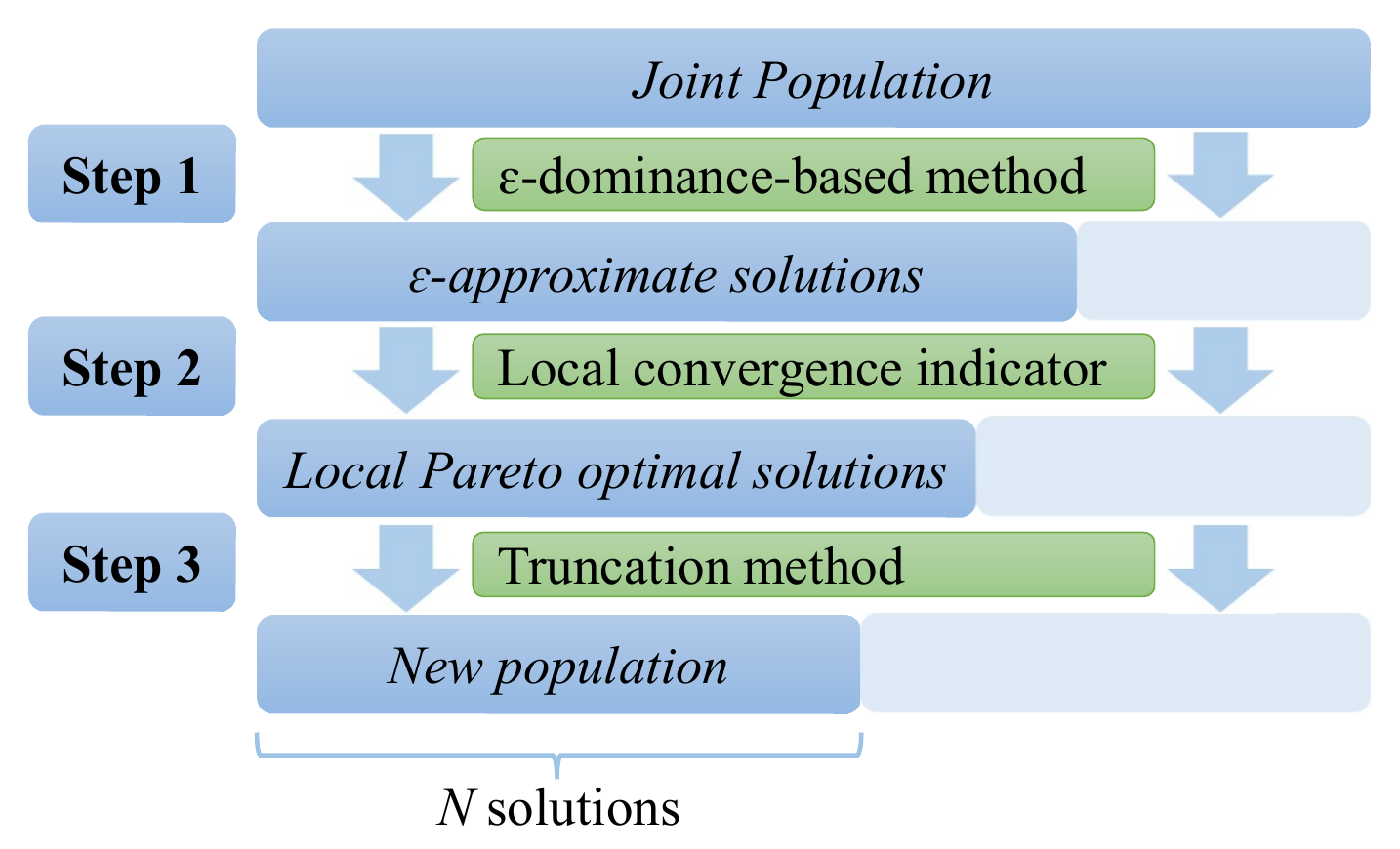}
\caption{Diagram of the environmental selection for diversity archive.}
\label{fig_selection}
\end{center}
\end{figure}

\begin{figure*}[tbh]
\begin{center}
\includegraphics[width=7in]{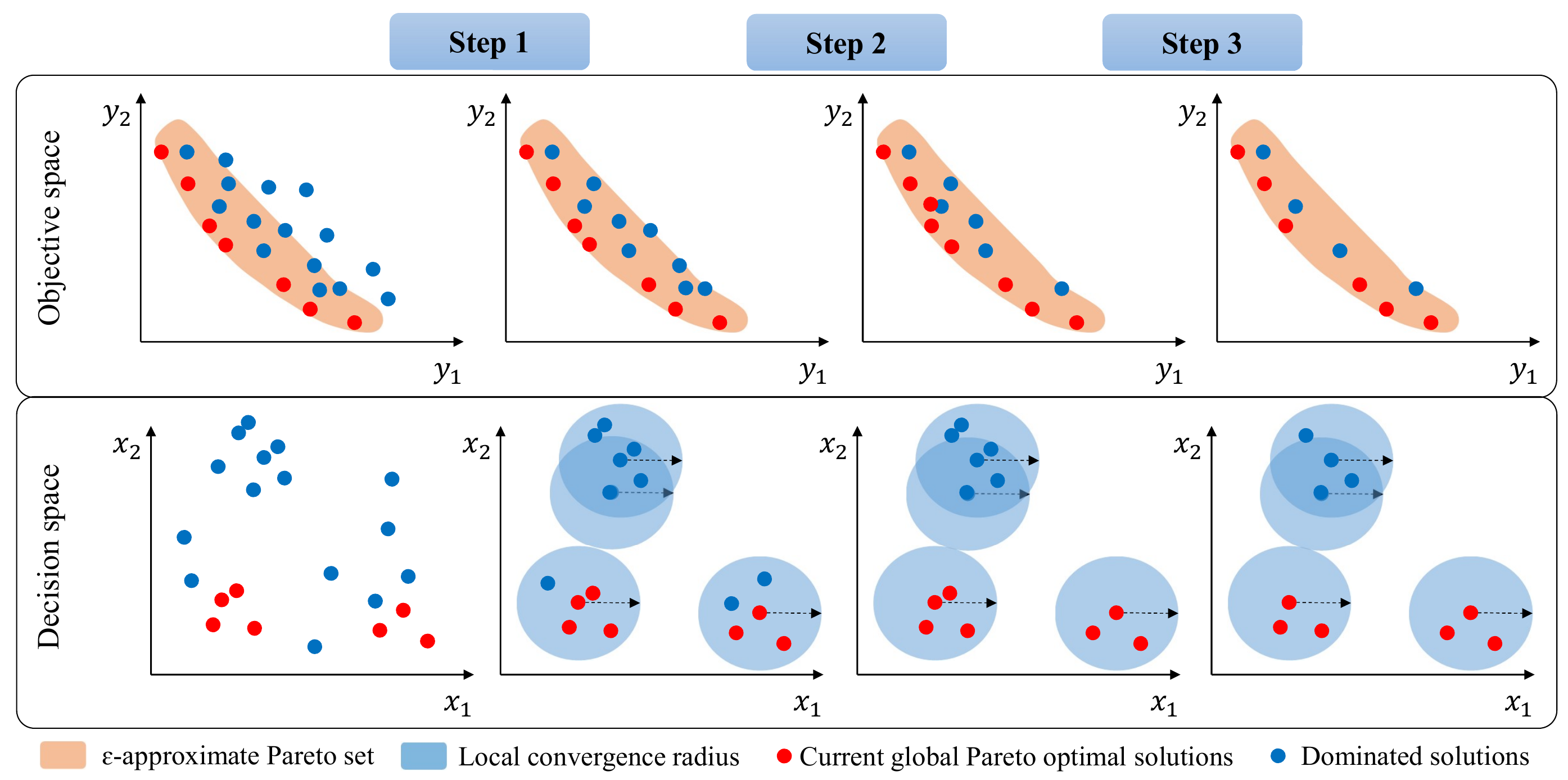}
\caption{Diagram of the solution selection process in the diversity archive.}
\label{fig_selection2}
\end{center}
\end{figure*}

\begin{algorithm}[htb]
\caption{Environmental selection for diversity archive}
\label{alg_update}
\begin{algorithmic}[1]
\REQUIRE {Diversity archive $DA$, offspring from convergence archive $OffC$, offspring from diversity archive $OffD$, population size $N$, evolution status $s \in [0,1]$, parameter for $\epsilon$-approximate PS $\epsilon$}
\ENSURE {Updated diversity archive $DA$, individual fitness $FitnessD$ }
\STATE $JointArc \leftarrow DA \cup OffC \cup OffD$
\STATE $FNo \leftarrow NDSort(JointArc)$ /* Obtain the Pareto rank of the JointArc. */
\STATE $GlobalPF \leftarrow JointArc(FNo==1)$ /* Obtain the global PF. */
\STATE $\epsilon _i \leftarrow CalEps(\epsilon,s)$ /* Calculating the current $\epsilon$ value according to \eref{equ_eps}. */
\STATE $EpsApxSol \leftarrow EpsNDSort(JointArc,GlobalPF,\epsilon _i)$ /* Obtaining the $\epsilon$-approximate PS. */
\STATE $I_{LC} \leftarrow CalLocalConvergence(EpsApxSol)$
\STATE $LocalNDSol \leftarrow EpsApxSol(I_{LC}==0)$ /* Obtain the local non-dominated solutions. */
\IF{$Num(LocalNDSol)<N$}
\STATE $IDX \leftarrow sort(I_{LC})$
\STATE $DA \leftarrow(EpsApxSol(IDX(1:N)))$
\ELSE
\STATE $DA \leftarrow Truncate(LocalNDSol,N)$
\ENDIF
\STATE $FitnessD \leftarrow CalCrowdDis(DA)$
\end{algorithmic}
\end{algorithm}

\fref{fig_selection} and \fref{fig_selection2} present the selection process of the $\epsilon$-dominance based method. In addition, the detailed operation can be seen from Algorithm \ref{alg_update}. In the beginning, the global PF is obtained through a fast non-dominated sorting method ($lines\ 1-5$), which is used to retain the $\epsilon$-approximate solutions, called \textbf{Step 1} in \fref{fig_selection} and \fref{fig_selection2}. After that, the local convergence indicator is calculated to select local non-dominated solutions ($lines\ 6-7$), called \textbf{Step 2}. Finally, a crowding distance based truncation method is performed to balance the convergence and diversity of solutions in both the decision and objective spaces, termed \textbf{Step 3}. The above-mentioned methods will be described in the following sub-sections in detail.

\subsubsection{$\epsilon$-dominance based method}
\begin{figure}[bh]
\begin{center}
\includegraphics[width=3in]{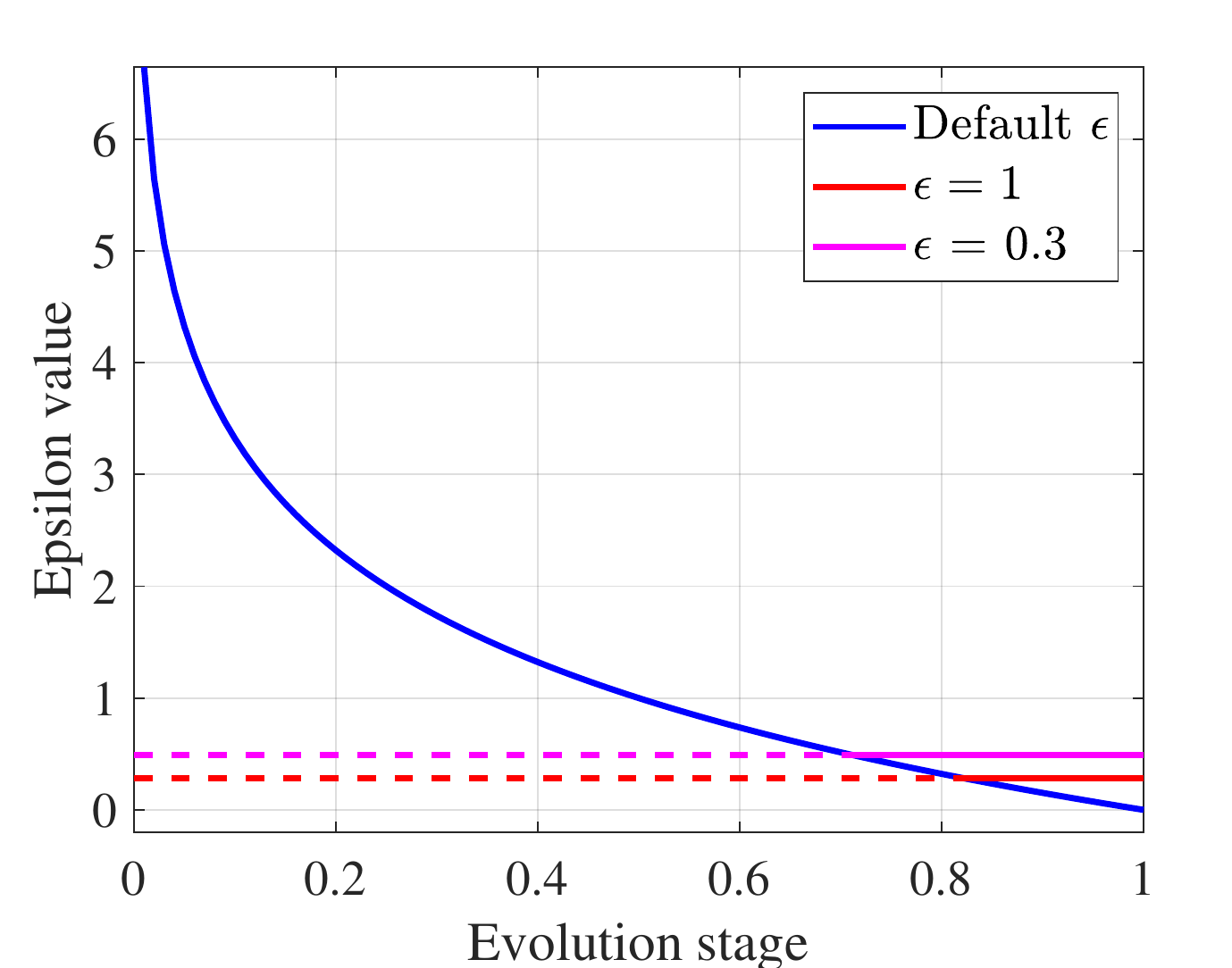}
\caption{Illustration of the change tendency of $\epsilon _i$ as the evolution stage changes under different $\epsilon$ settings.}
\label{fig_epschange}
\end{center}
\end{figure}

The use of $\epsilon$-dominance has two main purposes. The first is to avoid the case that solutions directly converge to the true PF, thus, enhancing the diversity of solutions. The second is to control the convergence of the local PSs. That is, in the early stage  $\epsilon$ is set to a large value to maintain the diversity. As the search progresses, the $\epsilon$ value decreases till it approaches a user-defined value. In particular, the setting of $\epsilon$ in the $i$-th generation is as follows:

\begin{equation}
\epsilon _i = \text{max} \{ log_2 (i/G), \epsilon \}
\label{equ_eps}
\end{equation}
where $G$ is the maximum generation. \fref{fig_epschange} displays the change tendency of $\epsilon _i$ with various  $\epsilon$ setting. As we can see, when $\epsilon$ is set to 1, then the value of $\epsilon _i$ remains constant throughout the last 50\% evolution. Notably, other functions that have the same tendency can also be used to adaptively adjust the value. In the following section, the effect of the parameter $\epsilon$ will be discussed in detail.

\subsubsection{Local convergence indicator}

Local convergence is a simple yet effective technique that has been applied in many multimodal optimization related studies. In \cite{li2022hierarchy}, Li et al. proposed a local convergence quality. Its calculation is shown as follows : 
\begin{equation}
c(i) = \frac{\sum_{j=1}^{n_i} B_{i,j}}{n_i}
\end{equation}
where $B_{i,j}$ is 1 if $\mathbf{x}_i$ is dominated by $\mathbf{x}_j$, and 0 otherwise; $n_i$ denotes the number of the neighbors for solution $\mathbf{x}_i$. The smaller $c(i)$, the better $\mathbf{x}_i$ is. 

The local convergence quality only considers the number of solutions that dominate itself. That is, two different solutions that non-dominated each other may have the same local convergence quality. Such situation usually occurs if the number of non-dominated solutions is small, which decreases the selection pressure substantially. We present an improved version based on the raw-fitness proposal in \cite{zitzler2001spea2}, termed local convergence indicator $I_{LC}$ \cite{li2022local}. Specifically, for each solution, both dominating and dominated solutions are taken into account. The calculation of $I_{LC}$ is as follows:

\begin{equation}
\label{equ_lc}
I_{LC}^i = \sum_{j \in \mathbf{N}_i} S_j \cdot D_{j, i}
\end{equation}
\begin{equation}
S_i = \sum_{j \in \mathbf{N}_i} D_{i,j}
\end{equation}
where $\mathbf{N}_i$ denotes the neighbor solution set of $\mathbf{x}_i$; $D_{i,j}=1$ refers to that $\mathbf{x}_i$ Pareto dominates $\mathbf{x}_j$. $S_i$ denotes the number of solutions it dominates. Importantly, $I_{LC}$ is to be minimized, i.e., $I_{LC}^i=0$ corresponds to a non-dominated solution, whereas a high $I_{LC}^i$ value means that $\mathbf{x}_i$ is dominated by a number of neighbor individuals.

To form stable niches to aid the calculation of $I_{LC}$, a distance-based method is applied. That is, solution $\mathbf{x}_j$ is taken as a neighbor solution of $\mathbf{x}_i$ only if their Euclidean distance is less than a threshold value $R$, see the following equation: 

\begin{equation}
R = \frac{\sum_{i,j=1}^{N} d_{i,j}}{2*N^2}
\end{equation}
where $d_{i,j}$ is the Euclidean distance between $\mathbf{x}_i$ and $\mathbf{x}_j$. $R$ represents half of the average distance between all solutions, which decreases gradually as the evolution goes on.

\begin{figure}[tbh]
\begin{center}
\includegraphics[width=3in]{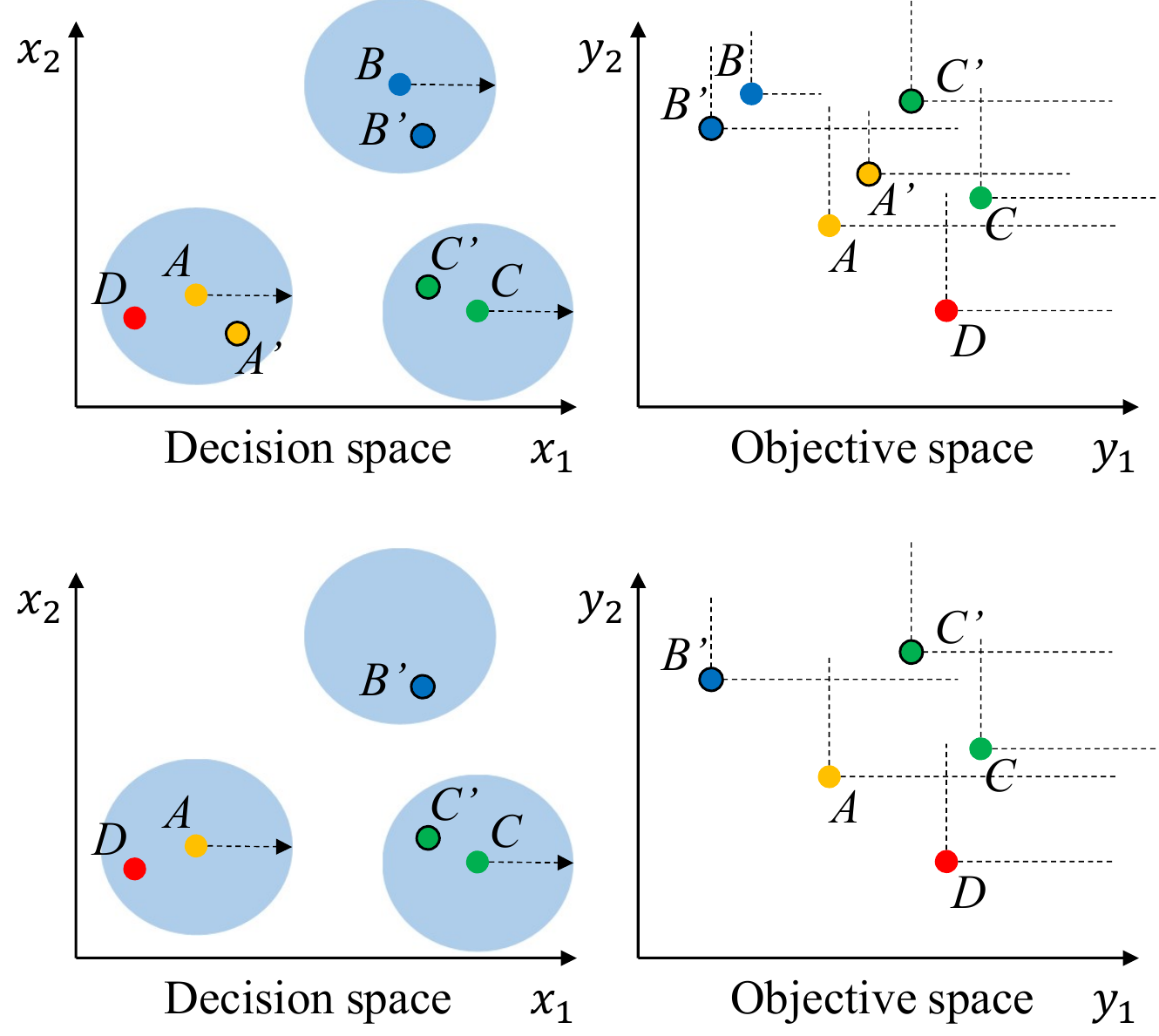}
\caption{Illustration of the local convergence indicator.}
\label{fig_localconvergence}
\end{center}
\end{figure}

\fref{fig_localconvergence} explains the diagram of $I_{LC}$. From the upper part, we can see that for $A$, $B$ and $C$, solutions inside the blue circles are considered neighbor solutions. Then, their $I_{LC}$ can be calculated. At the bottom of \fref{fig_localconvergence}, all local non-dominated solutions are retained, from which we can see that there are two global PSs (including solutions $B'$, $A$ and $D$) and one local PS (solutions $C$ and $C'$).

\subsubsection{Crowding distance based truncation method}

After obtaining local optimal $\epsilon$-approximate PS, we need to truncate the population size to $N$. The crowding distance based method as introduced in \cite{lin2020multimodal} is used:

\begin{equation}
\label{equ_dis}
CrowdDis_i = \frac{N-1}{\sum_{j=1}^{N}1/ \left \| \mathbf{x}_j-\mathbf{x}_i \right \| }
\end{equation}
where $\left \| \mathbf{x}_j-\mathbf{x}_i \right \|$ refers to the Euclidean distance of solutions $\mathbf{x}_i$ and $\mathbf{x}_j$. Notably, the decision vectors are normalized before the calculation.

To be more specific, while the number of the remaining solutions is larger than $N$, the solution with the maximum $CrowdDis$ will be discarded. This process will be iteratively executed until the population size is equal to $N$.

\section{Experiment}
\label{sec_exp}
\subsection{Experimental setting}
\label{sec_expset}
\subsubsection{Competitor algorithms}
To examine the performance of CoMMEA in dealing with GMMOPs, we have chosen seven state-of-the-art MMEAs as competitor algorithms, including MO\_Ring\_PSO\_SCD \cite{yue2017multiobjective}, DNEA-L \cite{liu2019searching}, MMODE\_CSCD \cite{liang2019multimodal}, CPDEA \cite{liu2019handling}, MMEA-WI \cite{li2021weighted}, MMOEA/DC \cite{lin2020multimodal} and HREA \cite{li2022hierarchy}. MO\_Ring\_PSO\_SCD is a representative algorithm in the field of multi-objective multimodal optimization; DNEA-L, MMOEA/DC and HREA are three representative MMEAs that perform well in searching for local PSs; MMODE\_CSCD, CPDEA and MMEA-WI are three new yet competitive MMEAs.

To ensure a fair algorithm comparison, for all algorithms, the population size $N$ is set to $100*D$, and the maximum number of function evaluations $N_{FE}$ is set to $5000*D$ when $D\leq 4$, where $D$ is the number of decision variables. In addition, except for MO\_Ring\_PSO\_SCD which uses the PSO operator, all the other algorithms adopt the simulated binary crossover (SBX) and polynomial mutation (PM) operators for generating offspring. The specific parameters in the comparison algorithms are set according to their original publications. For CoMMEA, the default value of $\epsilon$ is set to 0.1 and 0.3 for MMOPs and MMOPLs respectively, while we set $\epsilon=0.6$ for IDMP\_e test suite. All experiments are run on the PlatEMO \cite{tian2017platemo}, with a PC powered by AMD R9-5900X @ 3.70 GHz, 64G RAM. For the readers' convenience, the source code of CoMMEA is openly available. \footnote{Source code of CoMMEA and supplementary material can be found at \url{https://github.com/Wenhua-Li/CoMMEA}.}

\subsubsection{Benchmark}
To comprehensively examine the performance of CoMMEA, the test suites adopted here include the CEC2020 test suite, IDMP test suite, IDMP\_e test suite and multi-polygon test suite. CEC2020 test suite contains 22 problems. Some of them have both global and local PSs. IDMP includes 12 test problems with 2-4 decision variables and imbalanced searching difficulties. IDMP\_e contains only 2-3 decision variables, while the number of local PSs can be adjusted. The multi-polygon is a representative many-objective many-decision-variable MMOP test suite that has excellent visualization.

\subsubsection{Performance metrics}
Typically, the Inverted Generation Distance (IGD) and IGDX are used as performance metrics to comprehensively evaluate the approximation of the obtained solution sets to the true PF and true PS. For a solution set $\mathbf{X}$, the two metrics are calculated as follows:

\begin{equation}
IGD(\mathbf{X}) = \frac{1}{|\mathbf{X^*} |} \sum_{\mathbf{y} \in \mathbf{X^*}} 
\min_{\mathbf{x} \in \mathbf{X}} \{ED(F(\mathbf{x}),F(\mathbf{y})) \},
\end{equation}

\begin{equation}
IGDX(\mathbf{X})=\frac{1}{|\mathbf{X^*} |} \sum_{\mathbf{y} \in \mathbf{X^*}} 
\min_{\mathbf{x} \in \mathbf{X}} \{ED(\mathbf{x},\mathbf{y}) \}.
\end{equation}
where $ED(\mathbf{x},\mathbf{y})$ is the Euclidean distance between $\mathbf{x}$ and $\mathbf{y}$. $\mathbf{X}$ and $\mathbf{X^*}$ denote the obtained solution set and a set of a finite number of Pareto optimal solutions uniformly sampled from the true PS, respectively.

In addition, to evaluate the overall performance, the Friedman test is adopted. For each test problem, the 30 independent results of all MMEAs are used to calculate the ranks ($r_i^j$, where $i$ and $j$ are indexes of algorithms and test problems) by the Friedman test. Then, for a test suite, these ranks are summed to calculate the overall average ranks, which indicate the performances of MMEAs in the specific test suite.
\begin{equation}
R_i = \frac{\sum_{j=1}^{J}r_i^j}{J}
\end{equation}
where $J$ is the number of test problems, e.g., $J=12$ for IDMP test suite. The smaller the $R_i$, the better the $i$-th algorithm. 

\subsection{Results}
To provide a more intuitive comparison, the test suites are divided into four groups. Specifically, the first group contains part of the  CEC2020 test problems (MMF1-9, MMF 14, MMF1\_e, MMF1\_z and MMF14\_a), which is considered as low-dimension problems; the second group is IDMP test suite, which contains 12 test problems; the third group comprises the IDMP\_e test suite and part of CEC2020 test problems (MMF11-13, MMF 15, MMF15\_a and MMF16\_l1-l3), which contains 17 test problems and considered as the MMOPLs; the fourth group is multi-polygon, which contains 12 test problems. Overall, there are 13+12+17+10=54 test problems.

% Table generated by Excel2LaTeX from sheet 'total'
\begin{table*}[htbp]
\centering
\caption{The average Friedman ranks of all compared MMEAs on different test suites, where MO\_R, CSCD and WI are short names for MO\_Ring\_PSO\_SCD, MMODE\_CSCD and MMEA-WI respectively.}
%	\heavyrulewidth 0.1em
\begin{tabular}{cccccccccc}
\toprule[0.5mm]
{\textbf{Test Suites}} & {\textbf{Indicators}} & {\textbf{MO\_R}} & {\textbf{DNEA-L}} & {\textbf{CPDEA}} & {\textbf{CSCD}} & {\textbf{DC}} & {\textbf{WI}} & {\textbf{HREA}} & {\textbf{CoMMEA}} \\
\midrule
\multirow{2}[2]{*}{\textbf{CEC 2020}} & \textbf{IGD} & 5.95  & 6.95  & \textbf{2.20 } & 2.53  & 3.05  & 5.24  & 5.11  & 4.97  \\
& \textbf{IGDX} & 6.34  & 6.52  & 2.48  & 4.07  & 4.91  & 5.29  & 3.69  & \textbf{2.72 } \\
\midrule
\multirow{2}[2]{*}{\textbf{IDMP}} & \textbf{IGD} & 8.00  & 3.61  & \textbf{2.66 } & 3.09  & 6.56  & 5.00  & 2.92  & 4.16  \\
& \textbf{IGDX} & 6.79  & 5.27  & 4.61  & 5.92  & 3.54  & 4.27  & 3.88  & \textbf{1.72 } \\
\midrule
\multirow{2}[2]{*}{\textbf{MMOPLs}} & \textbf{IGD} & 6.59  & 4.20  & 6.18  & 5.85  & 2.13  & 7.09  & 2.33  & \textbf{1.63 } \\
& \textbf{IGDX} & 6.10  & 4.15  & 6.71  & 5.98  & 2.60  & 6.95  & 1.84  & \textbf{1.67 } \\
\midrule
\multirow{2}[2]{*}{\textbf{Multi Polygon}} & \textbf{IGD} & 6.80  & 2.75  & 8.00  & 3.78  & 5.50  & \textbf{1.51 } & 4.41  & 3.25  \\
& \textbf{IGDX} & 6.29  & 4.56  & 8.00  & 4.64  & 3.13  & 2.24  & 5.65  & \textbf{1.49 } \\
\midrule
\multirow{2}[2]{*}{\textbf{Overall}} & \textbf{IGD} & 6.80  & 4.41  & 4.84  & 3.98  & 4.09  & 4.94  & 3.59  & \textbf{3.36 } \\
& \textbf{IGDX} & 6.35  & 5.06  & 5.51  & 5.21  & 3.48  & 4.90  & 3.59  & \textbf{1.89 } \\
\bottomrule[0.5mm]
\end{tabular}%
\label{tab_result}%
\end{table*}%

\begin{figure}[tbh]
\begin{center}
\includegraphics[width=3.5in]{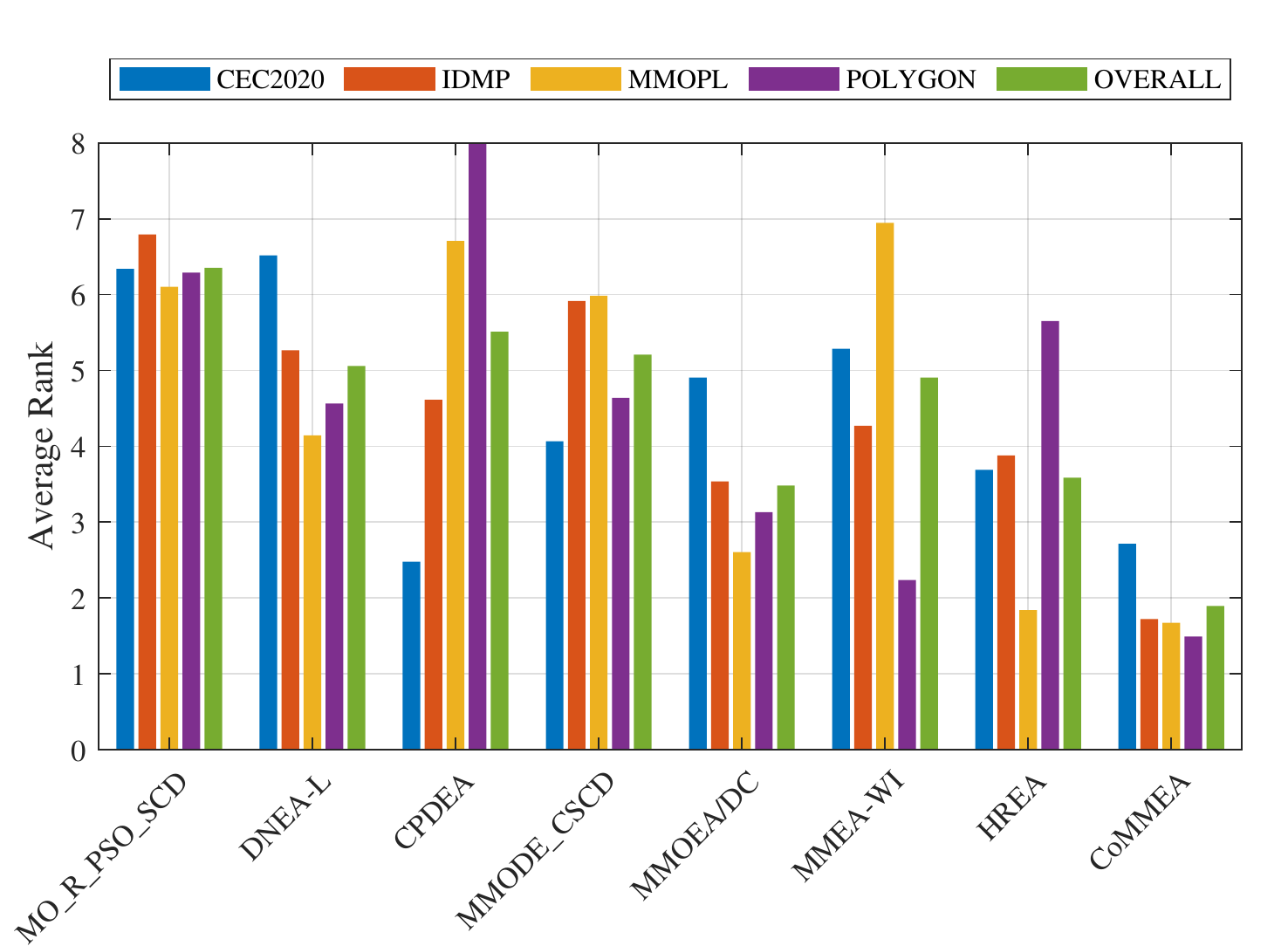}
\caption{Average ranks of all the competitor algorithms on the four group test problems in terms of IGDX metric.}
\label{fig_rank}
\end{center}
\end{figure}

\tref{tab_result} lists the overall average ranks of all the compared algorithms on four group test problems in terms of IGDX and IGD. Due to the page limitation, the detailed running results are provided in the supplementary material. In addition, \fref{fig_rank} intuitively shows the average ranks of all algorithms in terms of IGDX.

\subsubsection{Performance on CEC2020 test problems}
As we can see from \tref{tab_result}, for the CEC2020 test suite (13 problems in total), CPDEA performs the best in terms of IGD, while CoMMEA performs poorly. As for the detailed results presented in Table S-I, although CPDEA performs the best in terms of the average rank, MMODE\_CSCD wins 8 test problems over all MMEAs while CPDEA wins only 3 test problems. Since CoMMEA considers both global and local PSs and we set $\epsilon=0.1$ for these test problems, there is still room for performance improvement in terms of IGD. Weirdly, DNEA-L performs the least well on this test suite. DNEA-L is designed for MMOPLs, where the multi archive update method degrades the performance of MMOPs.

In terms of IGDX, CoMMEA achieves the best rank, followed by CPDEA. To be specific, CoMMEA and CPDEA win 6 and 3 test instances, respectively. In addition, HREA and MMODE\_CSCD show competitive performance. Although the average rank varies between different MMEAs, their performance in finding equivalent PSs is acceptable since all PSs can be found and the IGDX values are small enough. To sum up, the compared algorithms can receive good results on simple MMOPs with small-scale decision variables, while CoMMEA performs the best in terms of IGDX and CPDEA shows superior performance in terms of IGD.

\subsubsection{Performance on IDMP test problems}
The IDMP test suite is designed to better examine the diversity-maintaining ability of MMEAs. The difficulties in obtaining different PSs are different. Therefore, algorithms are more likely to converge to the easy-find PS. In terms of IGD, CPDEA and HREA are two competitive algorithms, while CoMMEA is in the middle stage. It is interesting that MMODE\_CSCD wins all 4 instances with two decision variables, while CPDEA and DNEA-L win all the instances with three and four decision variables, respectively. This indicates that, in the objective space, DNEA-L is competitive in solving many-objective problems, while MMODE\_CSCD and CPDEA show good ability in obtaining a well-converged and well-distributed solution set.

For IGDX, CoMMEA is the best algorithm (rank 1.72), while the second-best is MMOEA/DC (rank 3.54). On the other hand, MO\_Ring\_PSO\_SCD and MMODE\_CSCD perform poorly. MO\_Ring\_PSO\_SCD is a preliminary representative MMEA that does not consider imbalanced character. For them, only one global PS can be found in most of the algorithm runs. To be specific, CoMMEA wins 7 instances while MMOEA/DC wins 3 (IDMPM4T2-4) and HREA wins 2. It seems that MMOEA/DC performs better than CoMMEA on problems with 4 objectives. In conclusion, CoMMEA, MMOEA/DC, HREA, MMEA-WI and CPDEA can find multiple PSs for the IDMP test suite, while CoMMEA is the most stable and high-performance algorithm in terms of IGDX.

\subsubsection{Performance on MMOPLs}
In this section, almost all existing MMOPLs are selected as benchmark problems to verify the ability of MMEAs to search for global and local PSs. As we can see from \tref{tab_result}, different from other test suites, the average ranks for MMOPLs are highly consistent in terms of IGD and IGDX. That is, the referenced true PF for IGD calculation consists of both global and local PFs. As a result, if the algorithm cannot find global PSs, its performance would be poor in terms of either IGD or IGDX.

\begin{figure*}[htb]
\centering
\subfigure[MO\_Ring\_PSO\_SCD]{\includegraphics[width=1.7in]{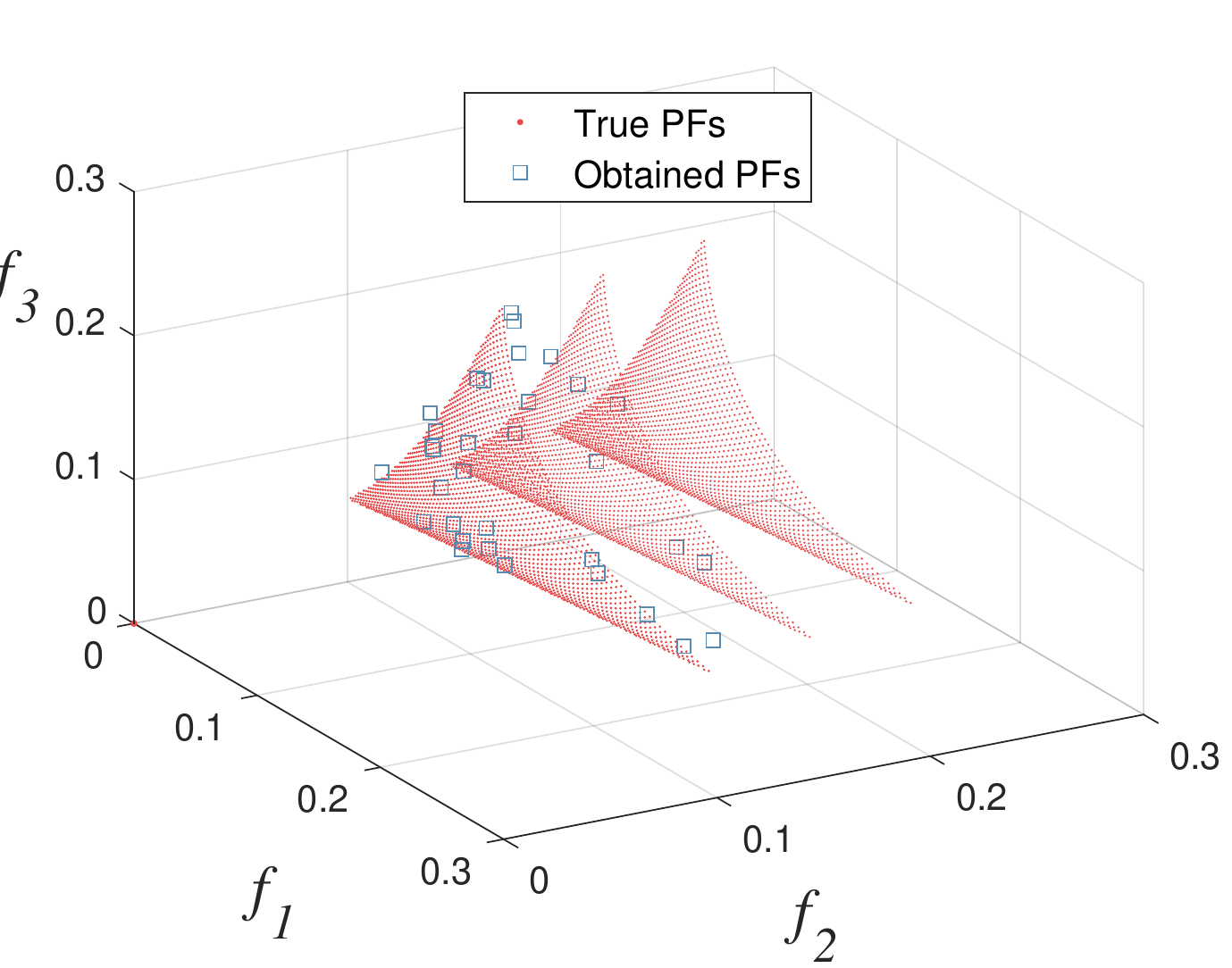}}
\subfigure[DNEA-L]{\includegraphics[width=1.7in]{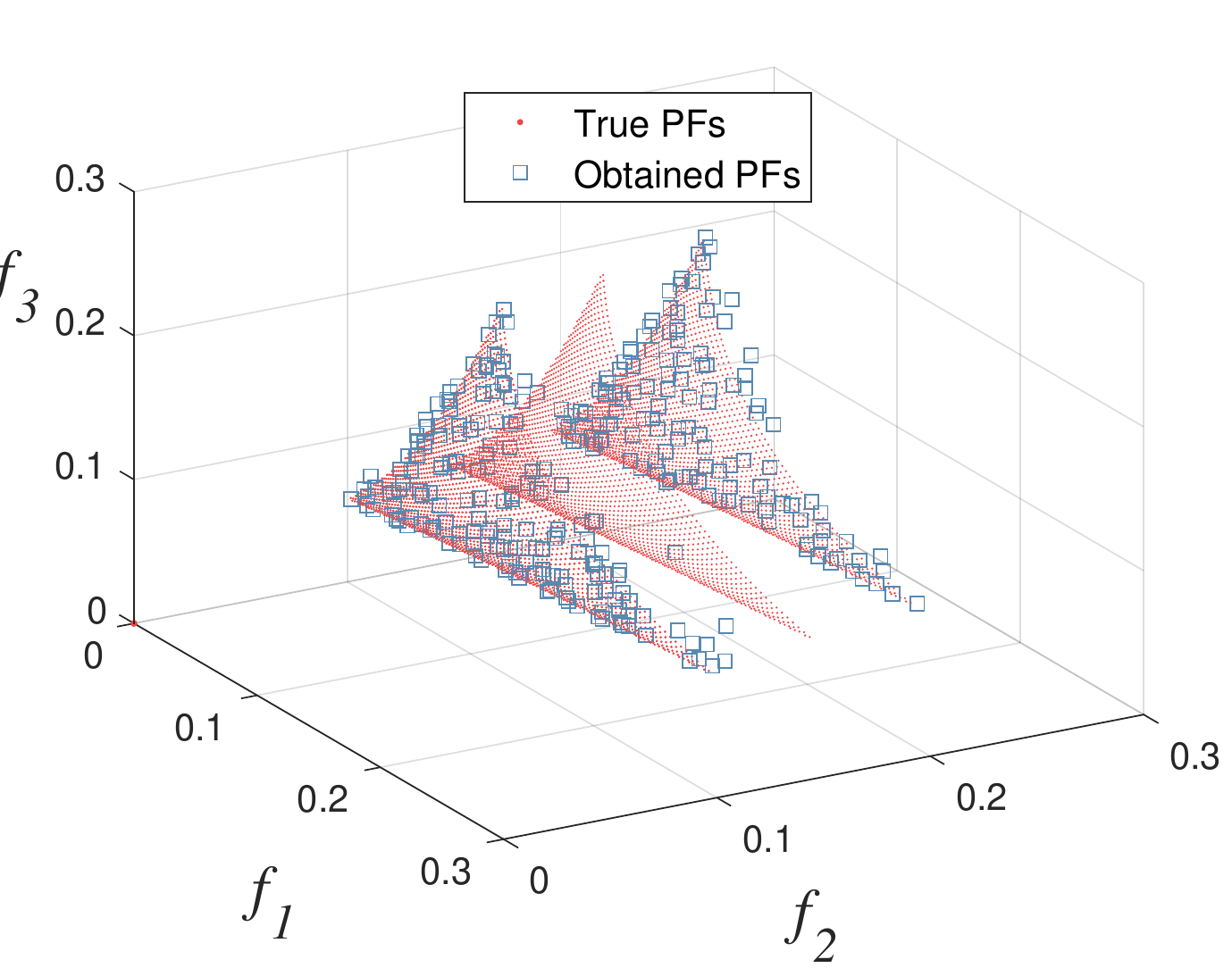}}
\subfigure[CPDEA]{\includegraphics[width=1.7in]{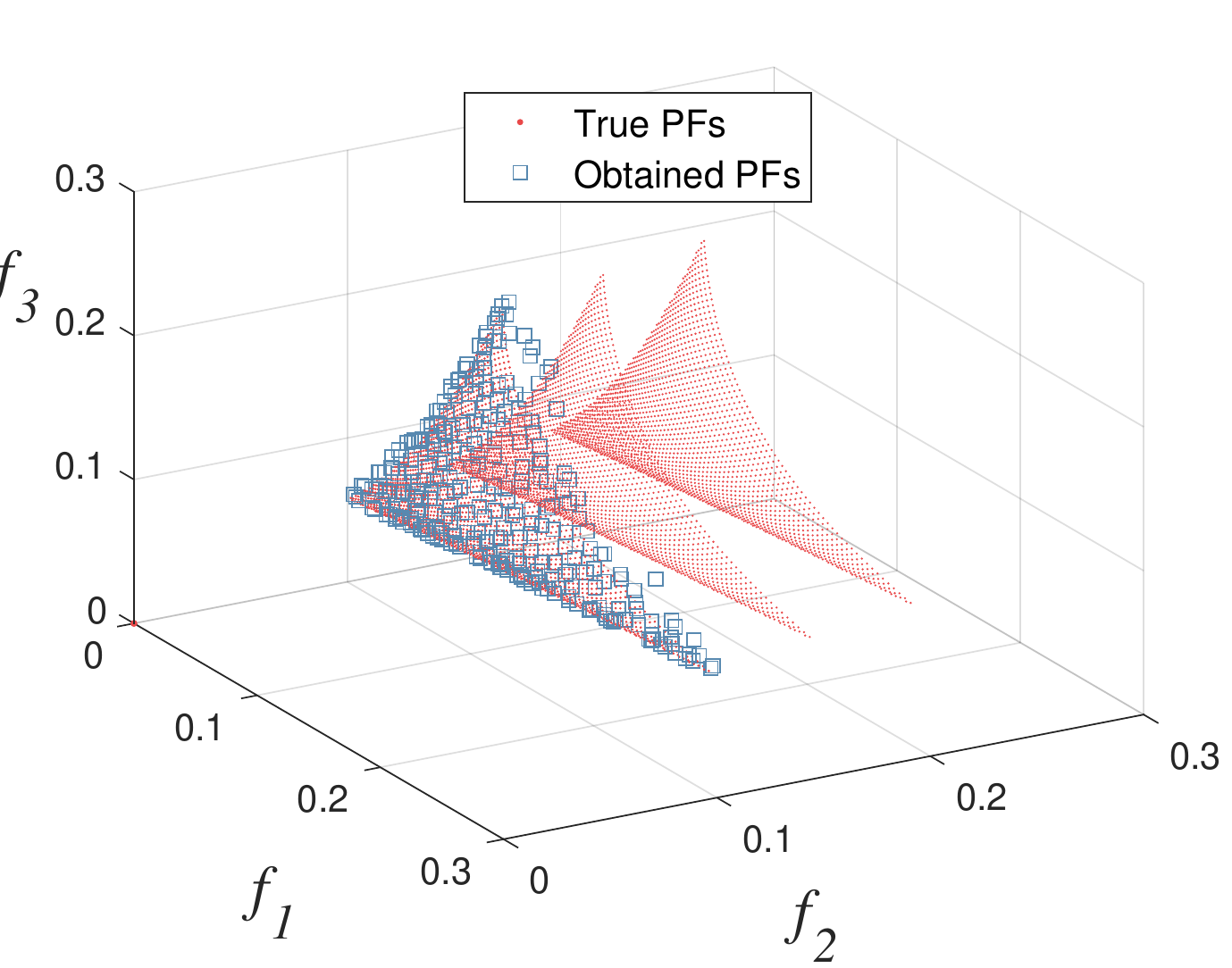}}
\subfigure[MMODE\_CSCD]{\includegraphics[width=1.7in]{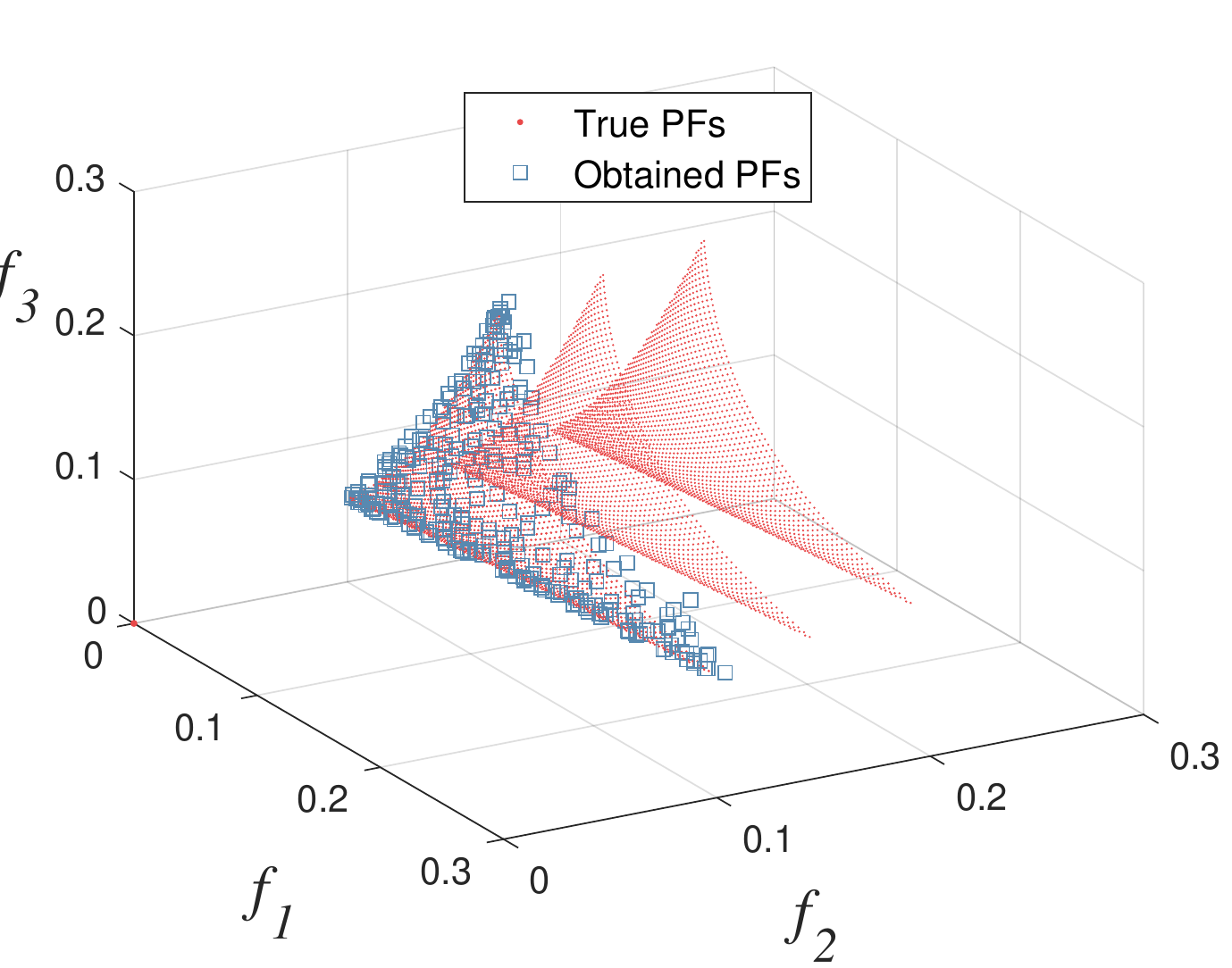}}
\subfigure[MMOEA/DC]{\includegraphics[width=1.7in]{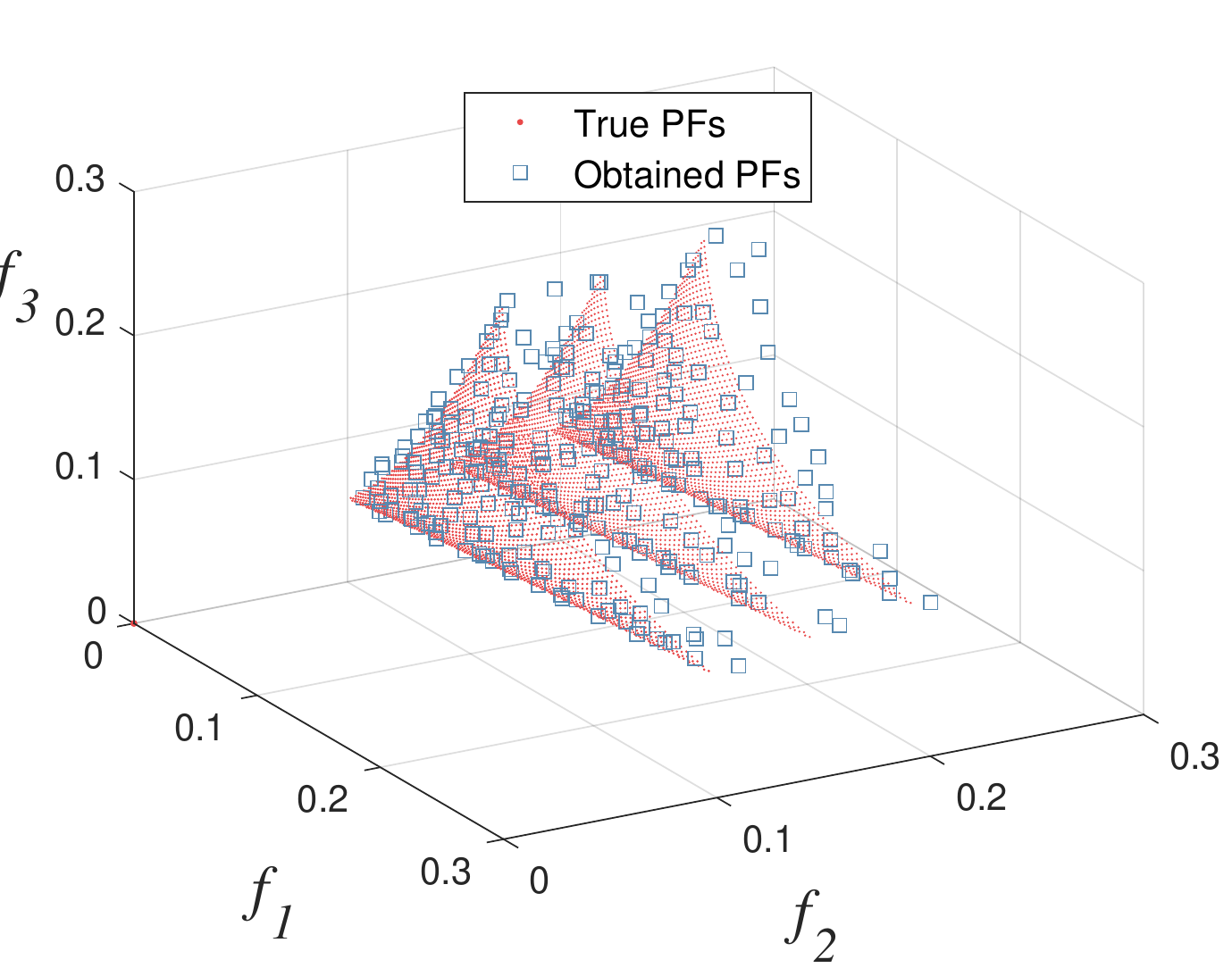}}
\subfigure[MMEA-WI]{\includegraphics[width=1.7in]{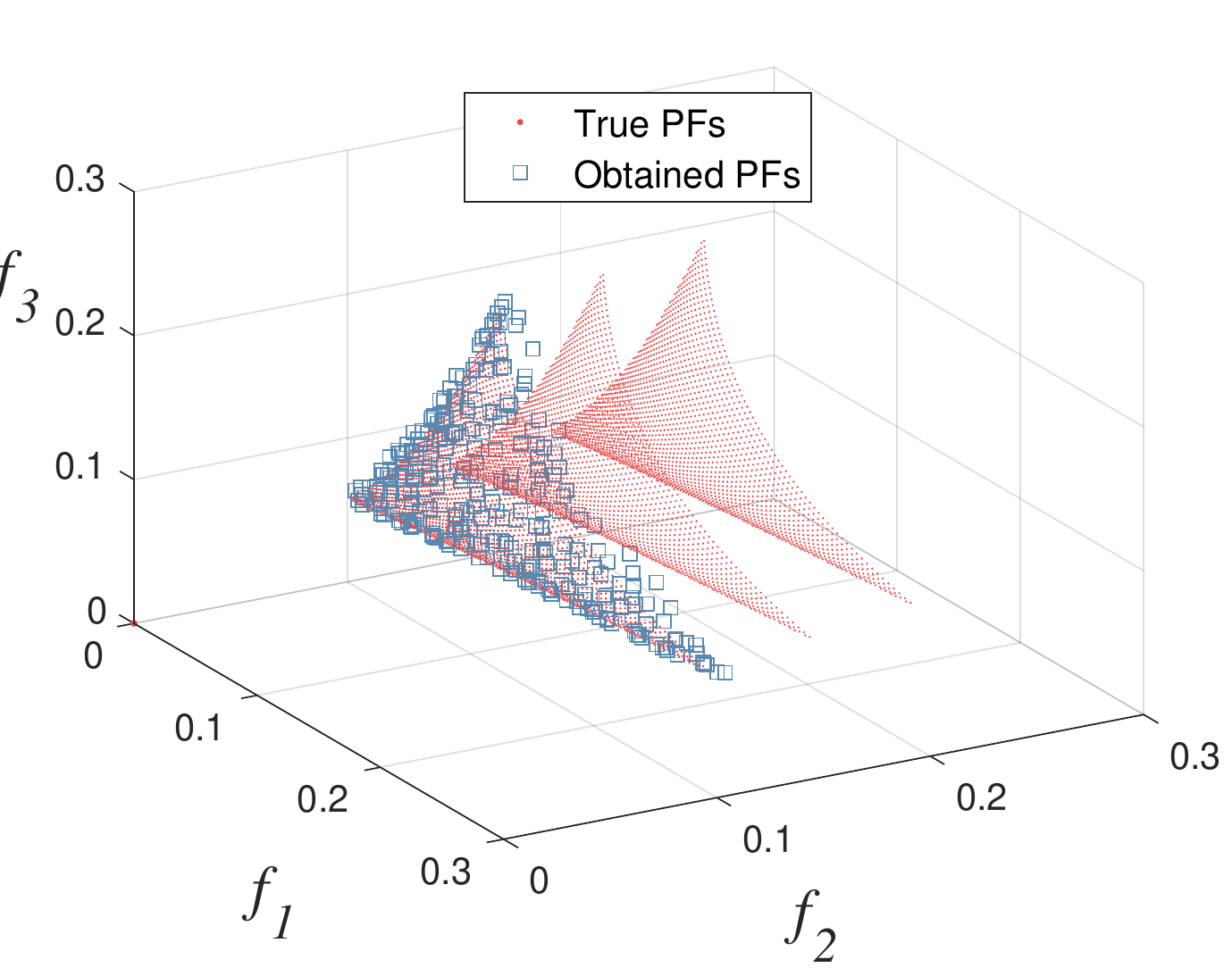}}
\subfigure[HREA]{\includegraphics[width=1.7in]{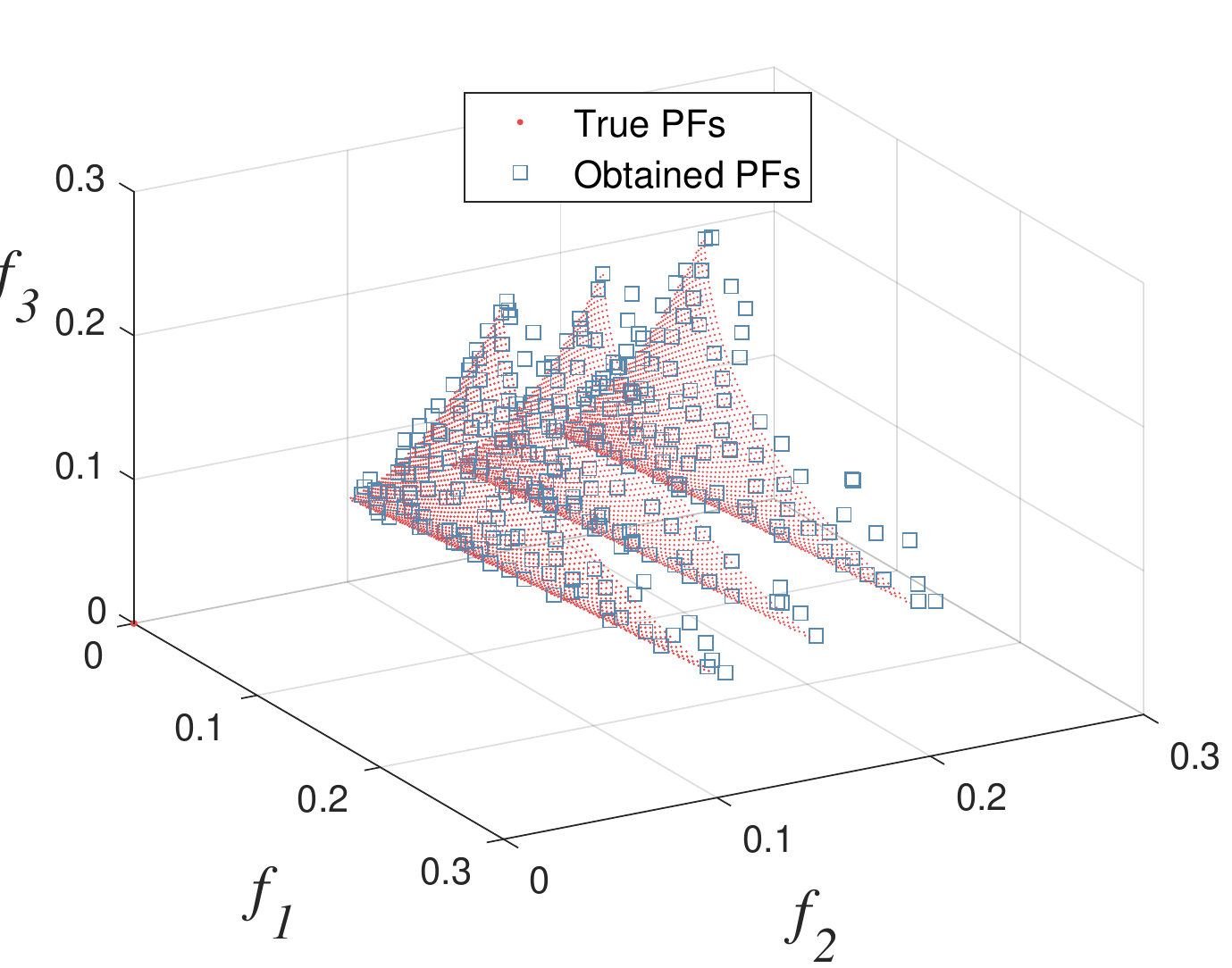}}
\subfigure[CoMMEA]{\includegraphics[width=1.7in]{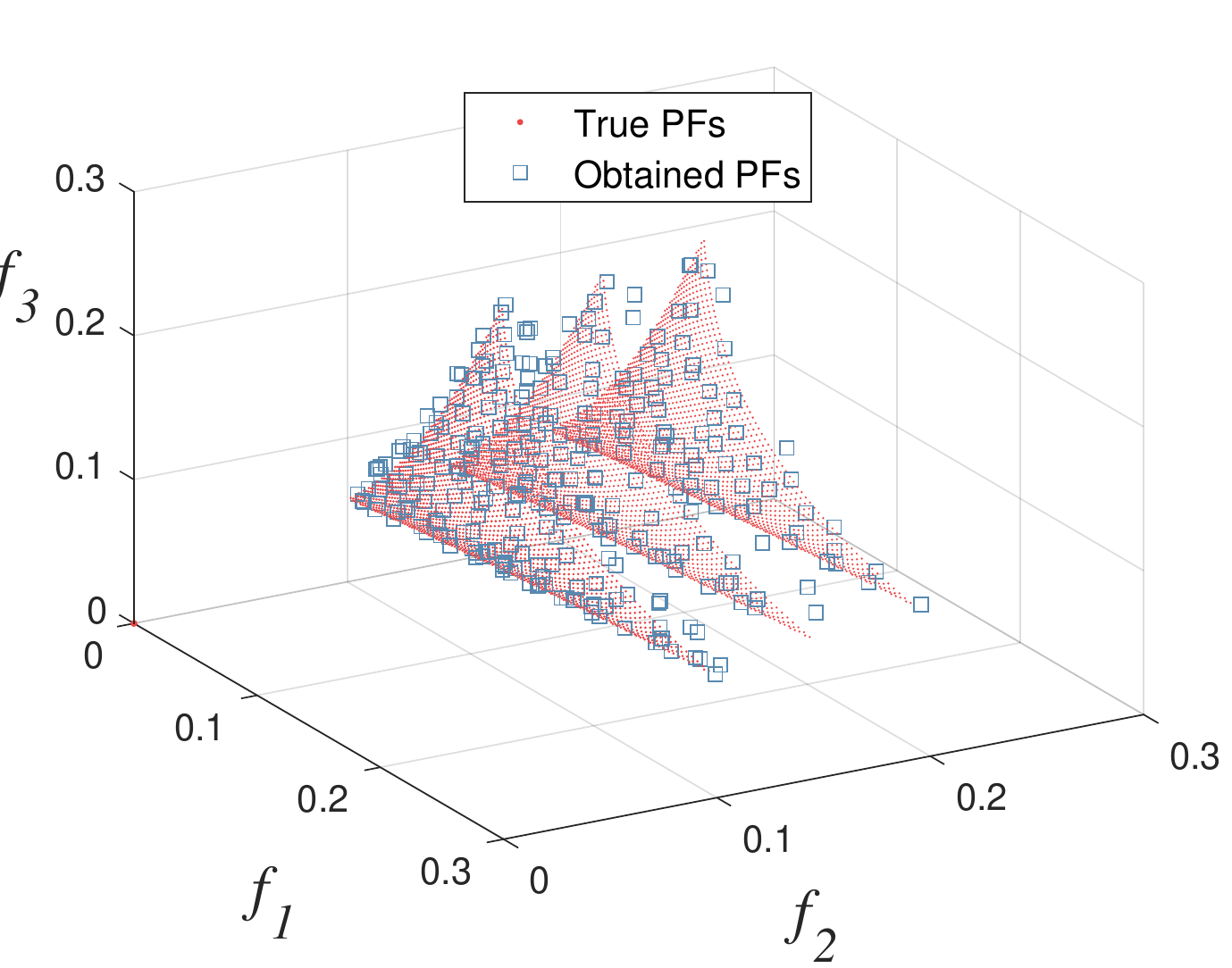}}

\caption{Distribution of solutions in the objective space on IDMPM3T2\_e obtained by different algorithms}
\label{fig_result}
\end{figure*}

CoMMEA, HREA and MMOEA/DC are the most competitive MMEAs for MMOPLs. As we can see from Table S-V and Table S-VII, they perform the best on 10, 5 and 2 in terms of IGDX and 7, 4 and 6 in terms of IGD, respectively. Although DNEA-L is designed specifically for MMOPLs, its performance is somehow poor. However, it still shows overwhelming superiority compared to other MMEAs that focus on only global PSs, e.g., MO\_Ring\_PSO\_SCD, CPDEA, MMODE\_CSCD and MMEA-WI. To intuitively present the results obtained by all compared algorithms, \fref{fig_result} presents solution distributions on IDMPM3T2\_e. It's worth mentioning that, the results closest to the average IGDX value are shown. As is presented, CoMMEA, HREA and MMOEA/DC can obtain all the global PF and local PFs, while DNEA-L can obtain part of the local PFs. Solutions obtained by the CoMMEA are closer to the true global and local PFs, which attributes to the introduction of the convergence archive. To sum up, CoMMEA is the best MMEA for solving existing MMOPLs.

\subsubsection{Performance on high-dimension problems}
Since the primitive MMEAs place little emphasis on MMOPs with many decision variables, their performances on such problems have not been well studied. In this section, we take the multi-polygon test suite as test problems and set the number of variables to 4, 8, 10, 14, 20 and 30 to examine the performance of CoMMEA. The overall results can be seen from \tref{tab_result} and \fref{fig_rank}.

\begin{figure*}[htb]
\centering
\subfigure[MO\_Ring\_PSO\_SCD]{\includegraphics[width=1.7in]{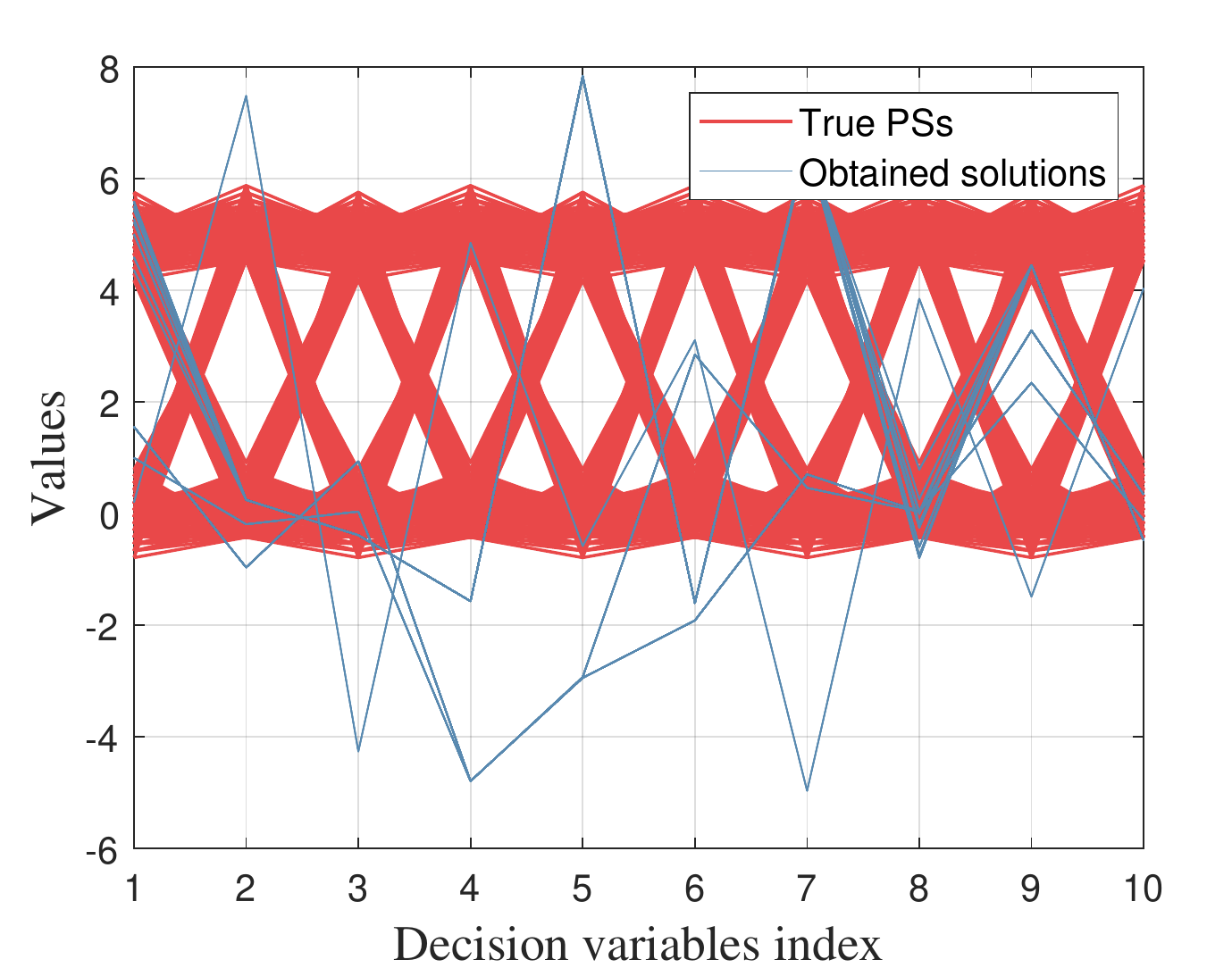}}
\subfigure[DNEA-L]{\includegraphics[width=1.7in]{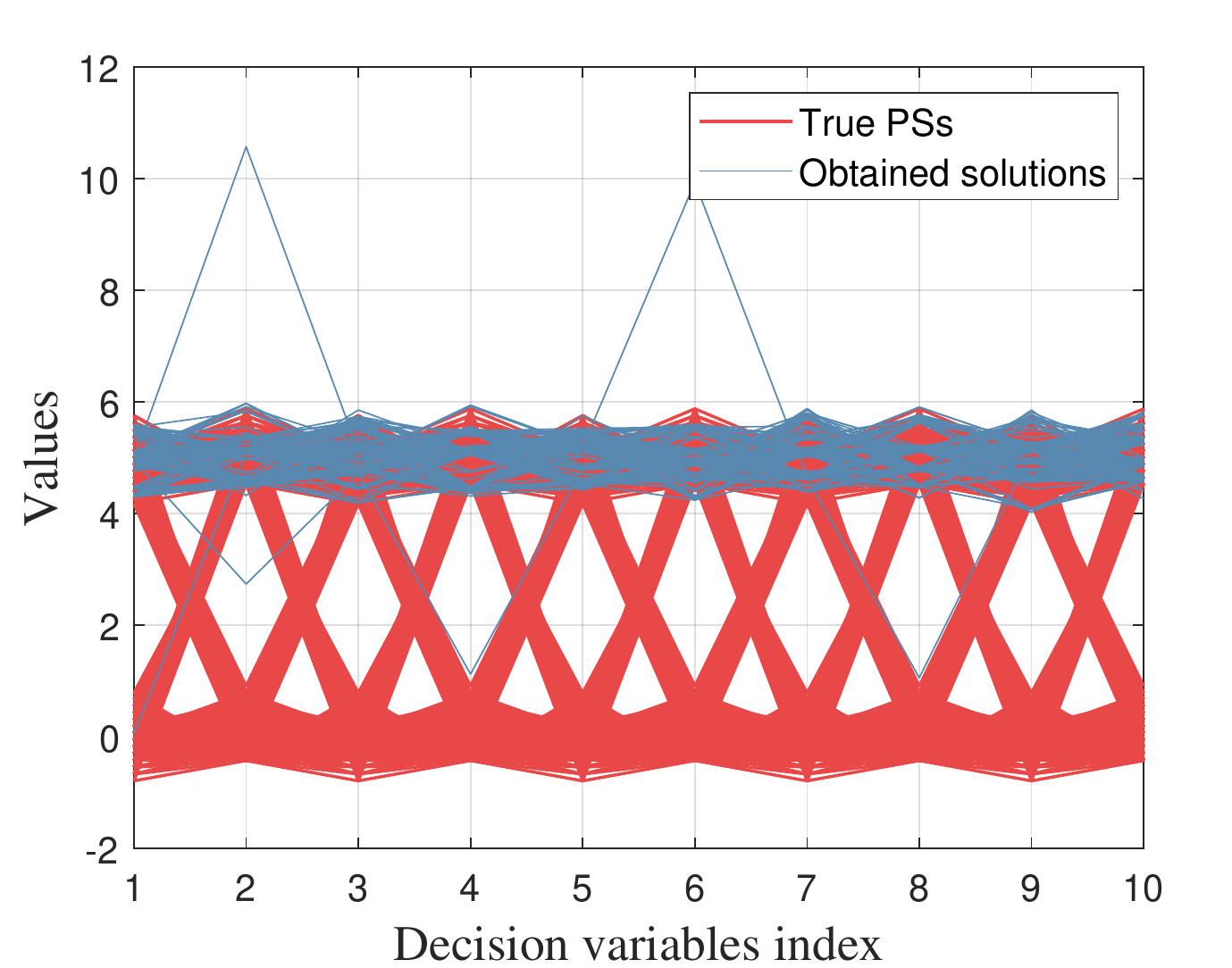}}
\subfigure[CPDEA]{\includegraphics[width=1.7in]{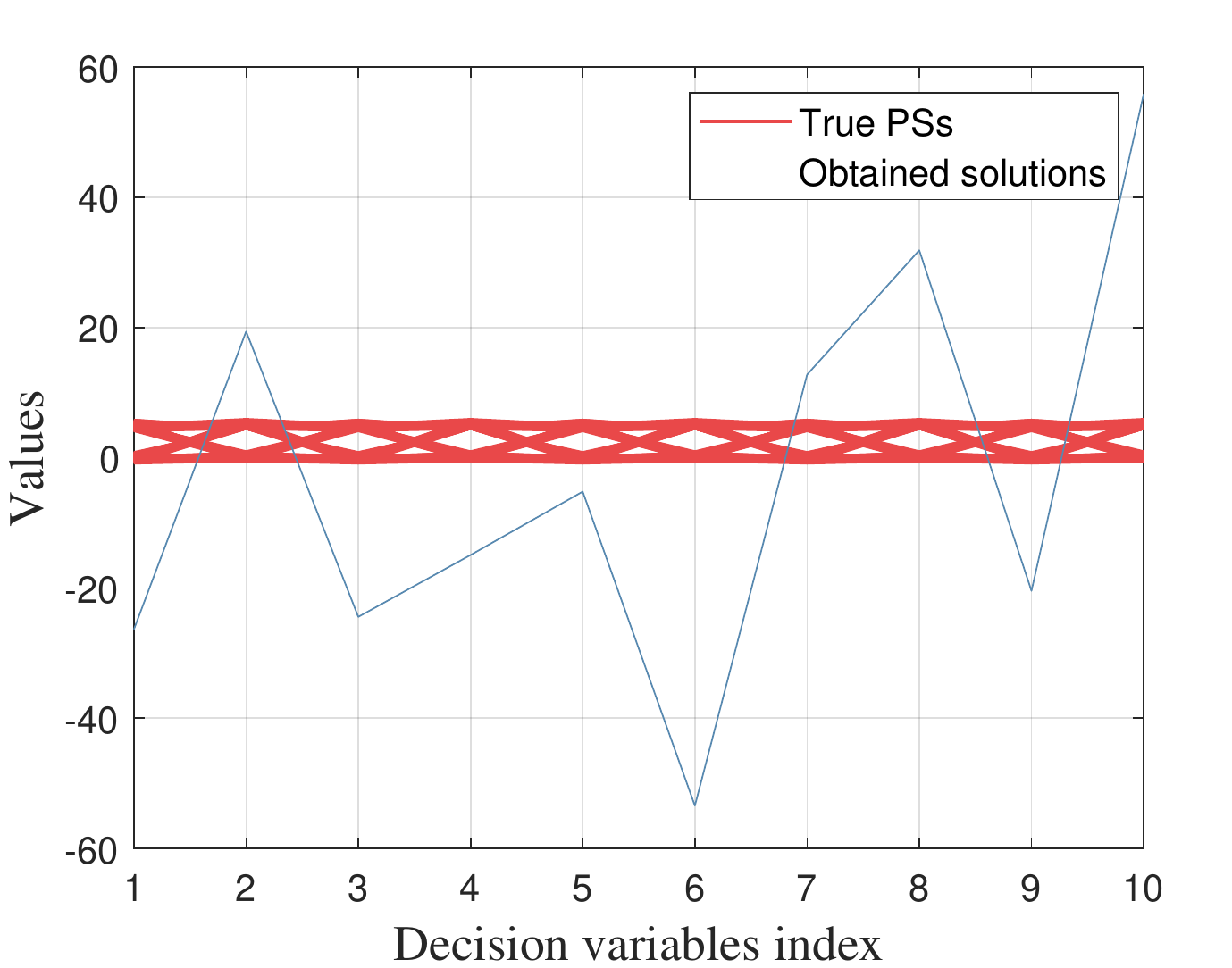}}
\subfigure[MMODE\_CSCD]{\includegraphics[width=1.7in]{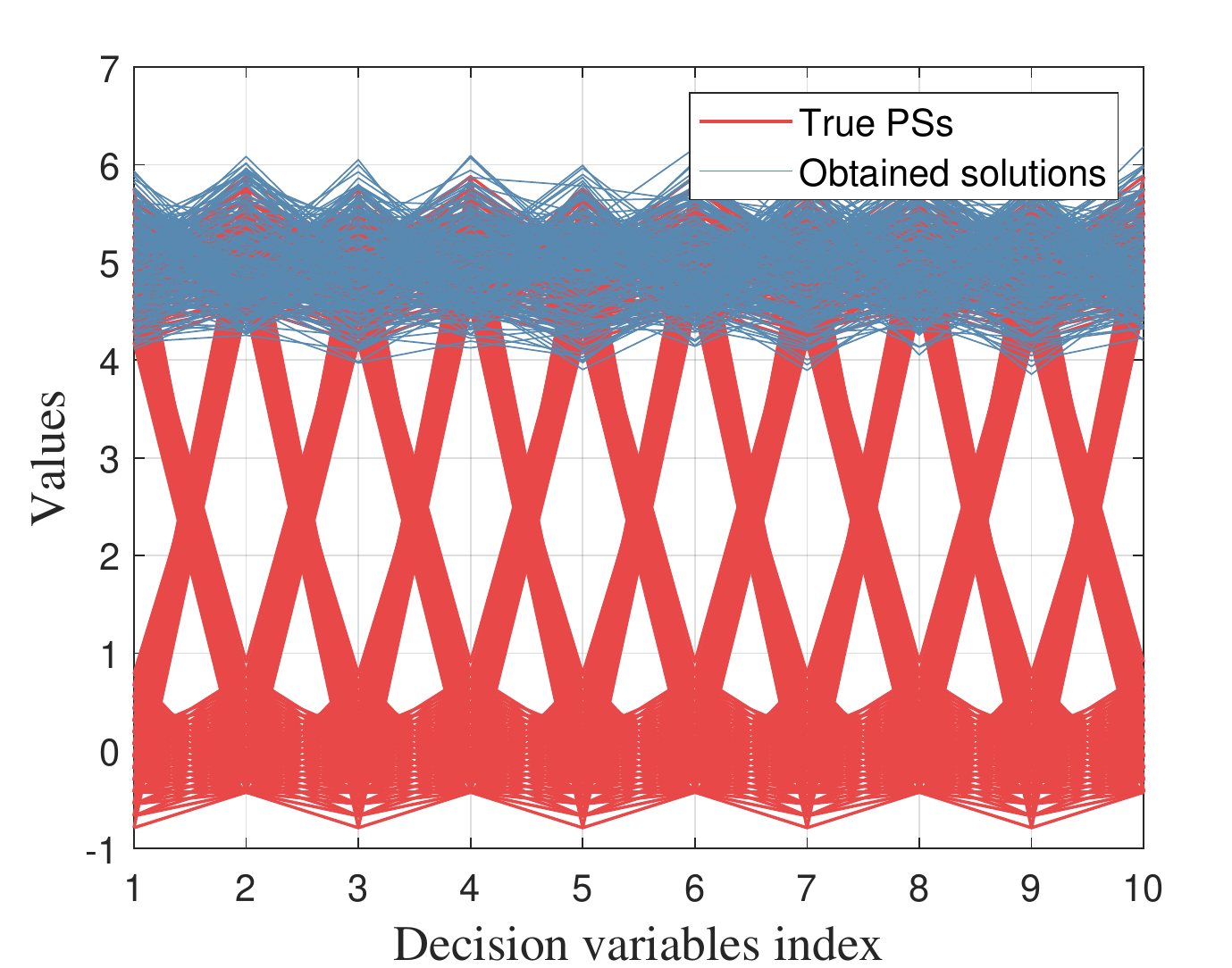}}
\subfigure[MMOEA/DC]{\includegraphics[width=1.7in]{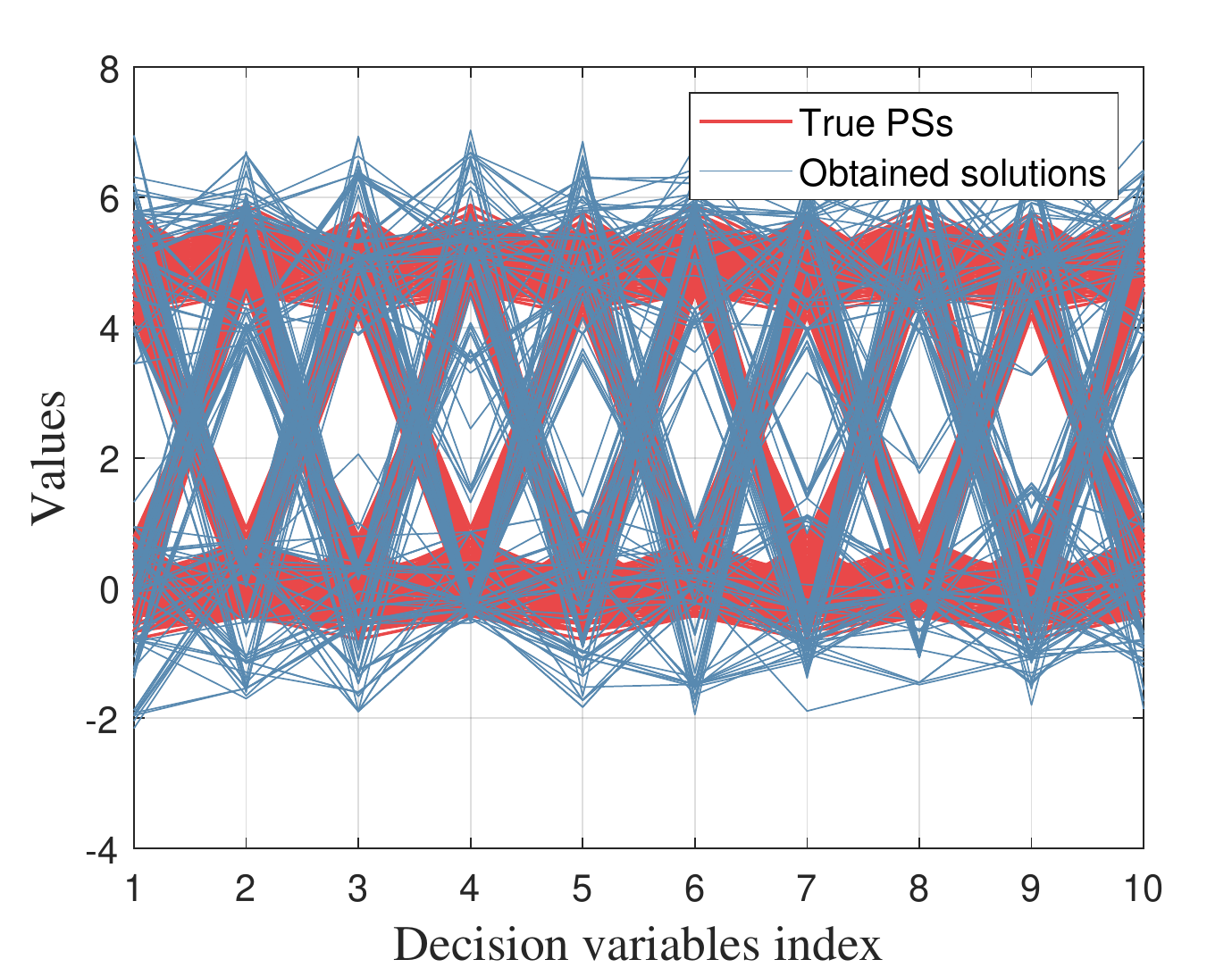}}
\subfigure[MMEA-WI]{\includegraphics[width=1.7in]{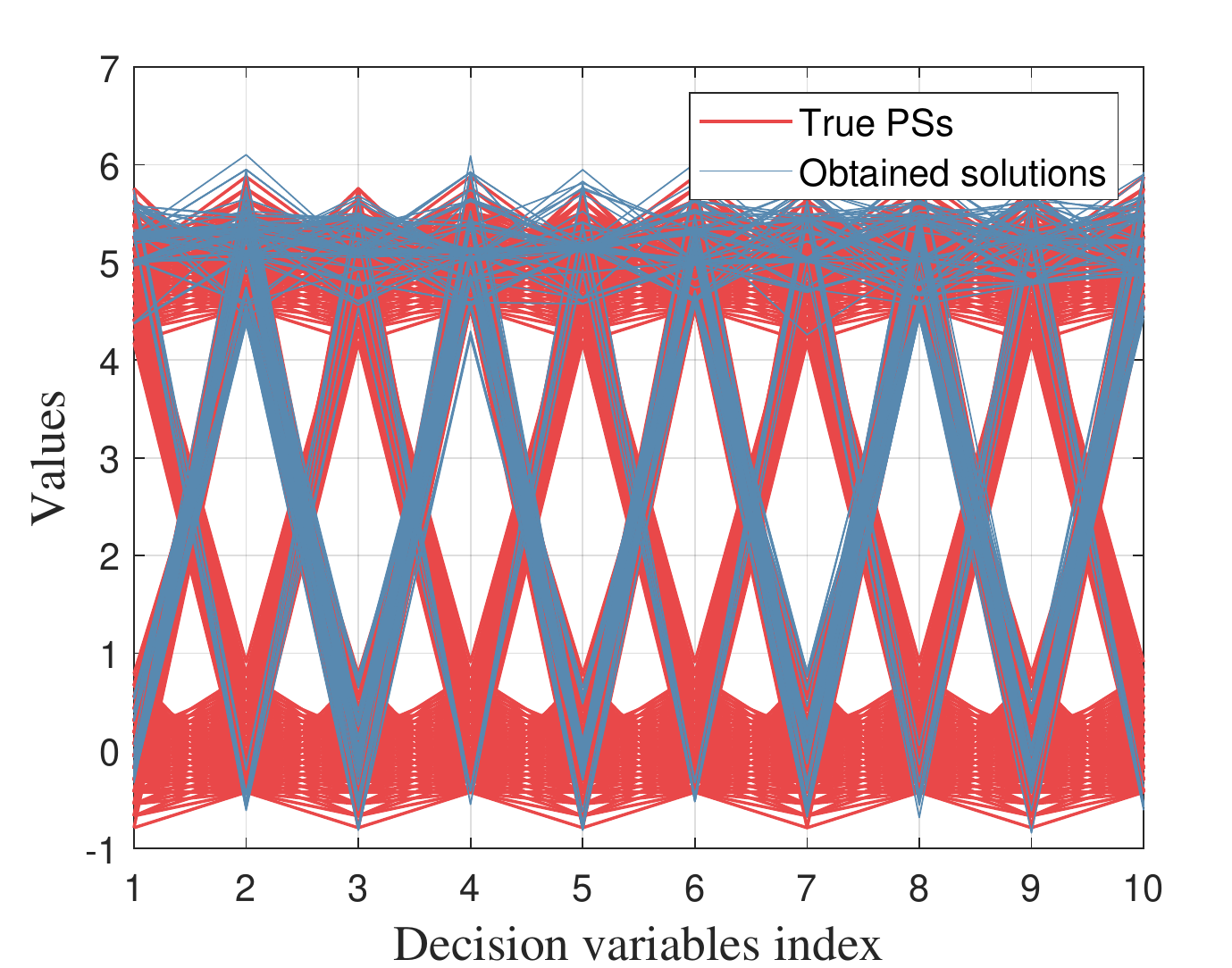}}
\subfigure[HREA]{\includegraphics[width=1.7in]{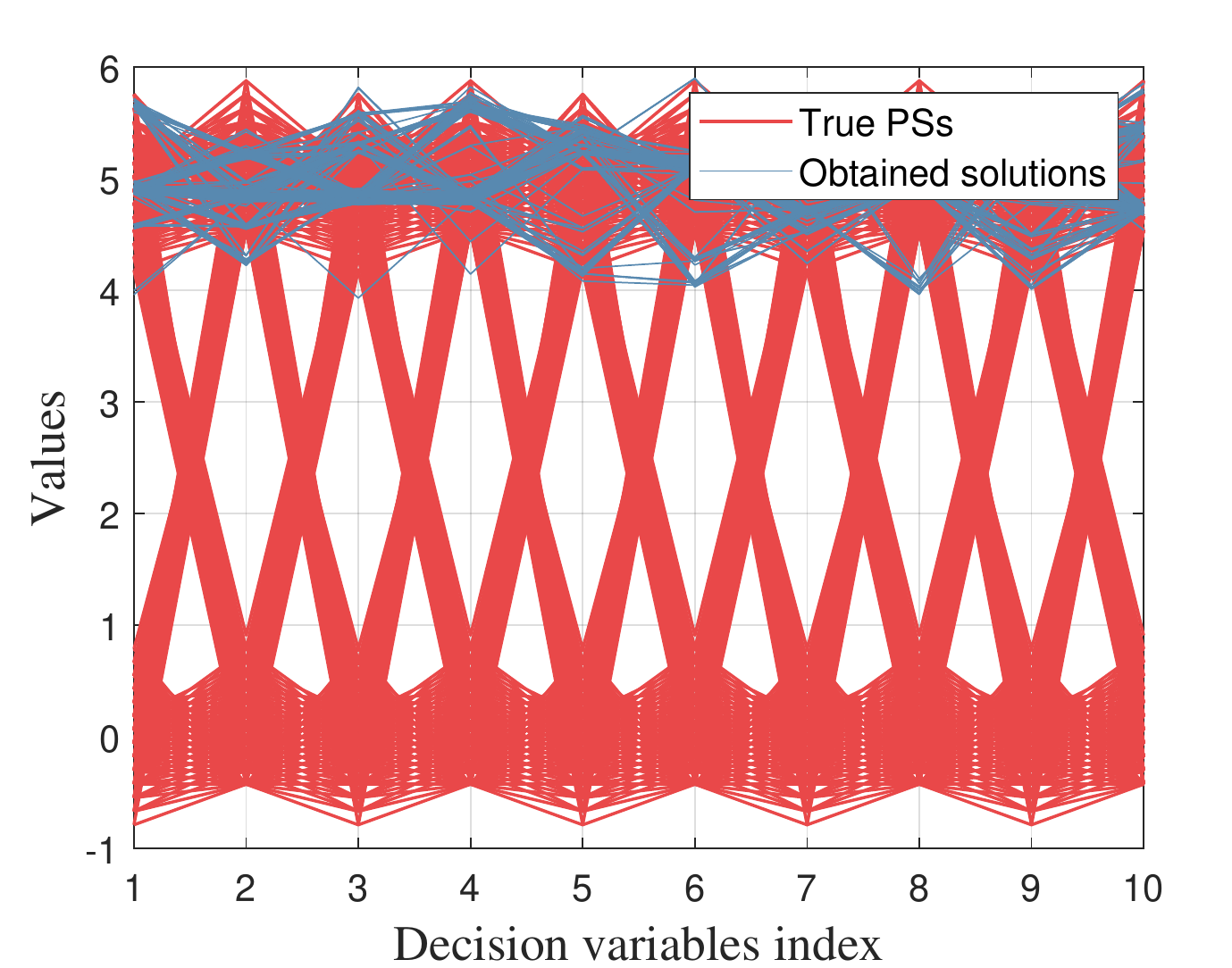}}
\subfigure[CoMMEA]{\includegraphics[width=1.7in]{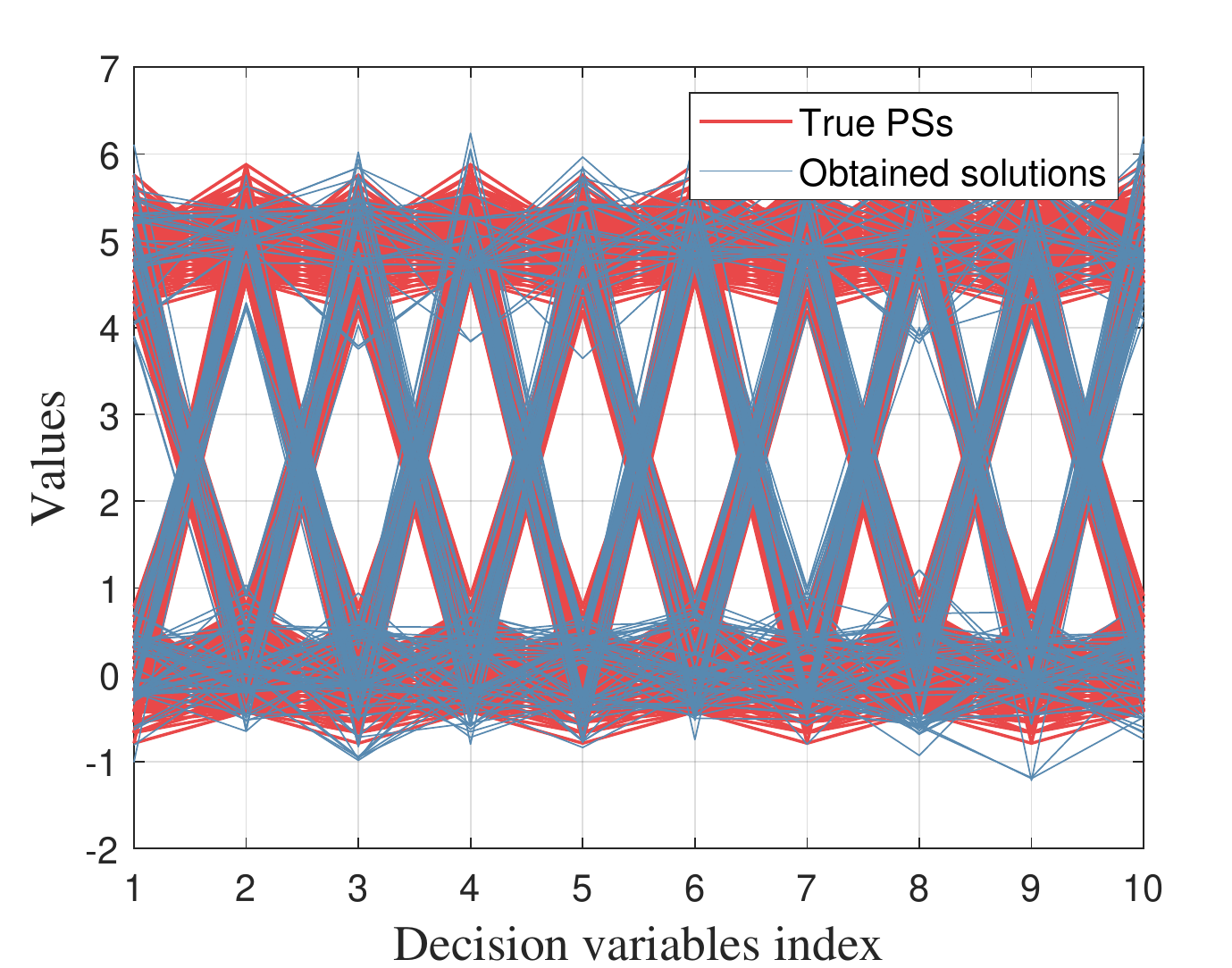}}

\caption{Distribution of solutions in the decision space on multi-polygon with 3 objectives and 10 decision variables  obtained by different algorithms}
\label{fig_result2}
\end{figure*}

As can be seen from the results, CoMMEA and MMEA-WI are the two best algorithms. To be specific, MMEA-WI performs the best in terms of IGD and wins 6 test problems. It is interesting that when the numbers of decision variables are 4 and 8, MMODE\_CSCD performs well in terms of IGD. In terms of IGDX, CoMMEA gets the best overall results and wins 9 test problems, where MMEA-WI and MMODE\_CSCD win 2 and 1 respectively. Again, MMODE\_CSCD shows good performance when there are only 4 decision variables. For problems with up to 30 decision variables, MMEA-WI performs a bit better than CoMMEA.

\fref{fig_result2} presents the distribution of the obtained solutions in the decision space (the most representative results are selected). As we can see, MO\_Ring\_PSO\_SCD and CPDEA cannot even converge to the true PS, while DNEA-L, MMODE\_CSCD and HREA can find only one equivalent PS.  MMOEA/DC, MMEA-WI and CoMMEA can find multiple PSs. However, the results obtained by MMOEA/DC contain many solutions that have poor convergence quality, which will cause difficulty for DMs. MMEA-WI cannot stably find all PSs. As a comparison, solutions obtained by CoMMEA have splendid convergence and distribution quality.

\subsubsection{Overall performance}
As indicated in \tref{tab_result} and \fref{fig_rank}, in terms of IGDX, CoMMEA performs the best on all the four group test problems. In addition, for all the 54 test problems, CoMMEA ranks 1.89 while the second best algorithm (MMOEA/DC) ranks 3.48, which indicates that CoMMEA has superior competitiveness in terms of IGDX. On the other hand, CoMMEA receives the best rank in terms of IGD for all the 54 test problems and the MMOPLs. The experimental results show that CoMMEA is a competitive MMEA for solving GMMOPs.

\section{Further Discussion}
\label{sec_discussion}
\subsection{Analysis on parameter $\epsilon$}
On the use of $\epsilon$-dominance, we have introduced a user-defined parameter $\epsilon$ to control the convergence quality of local PSs. To analyze the effect of $\epsilon$, we choose IDMPM2T4\_e as the test problem. Specifically, there are one global PF and two local PFs. Other parameters are set to the same as that in \sref{sec_expset}.

\begin{figure*}[htb]
\centering
\subfigure[$\epsilon$=0]{\includegraphics[width=2.2in]{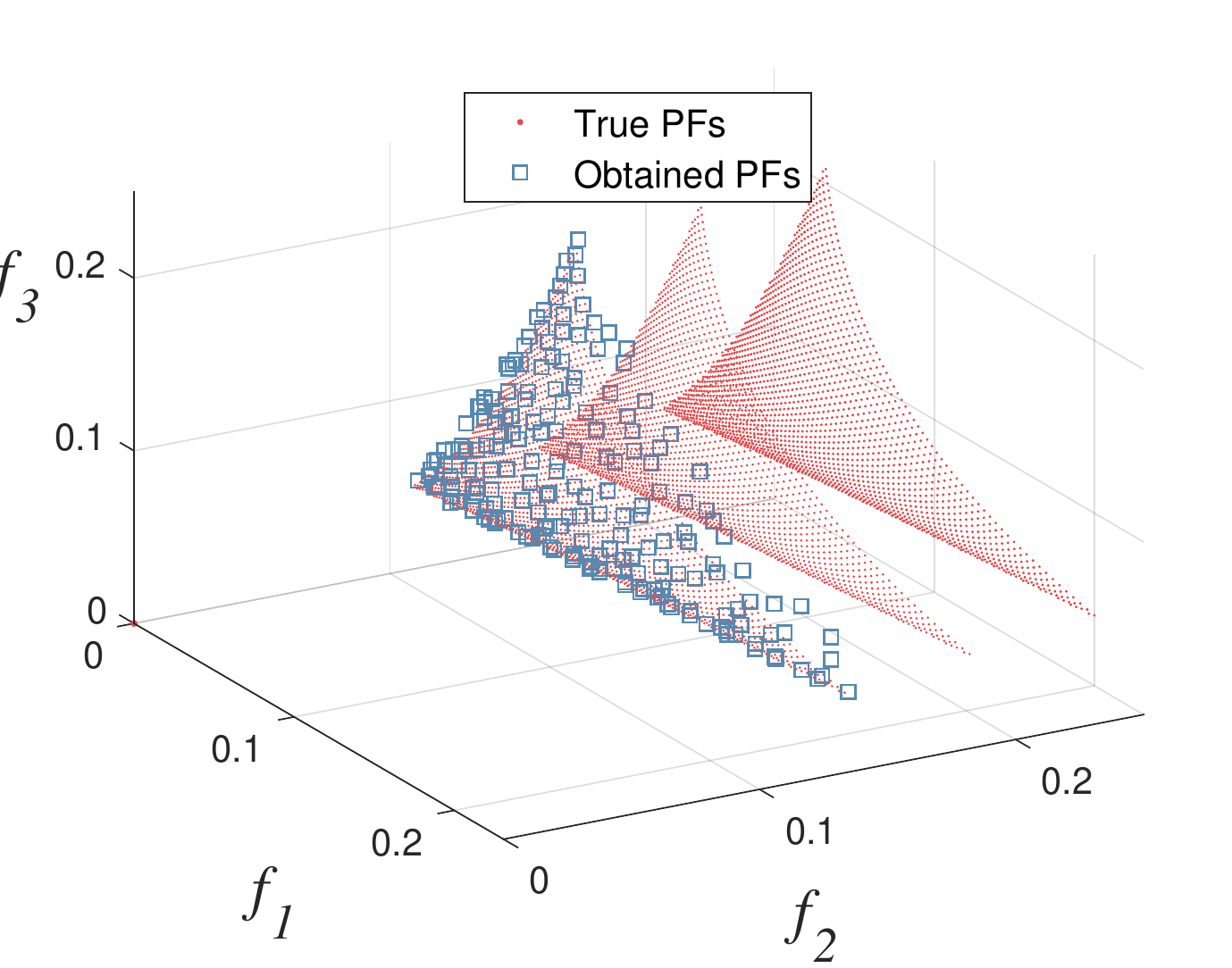}}
\subfigure[$\epsilon$=0.3]{\includegraphics[width=2.2in]{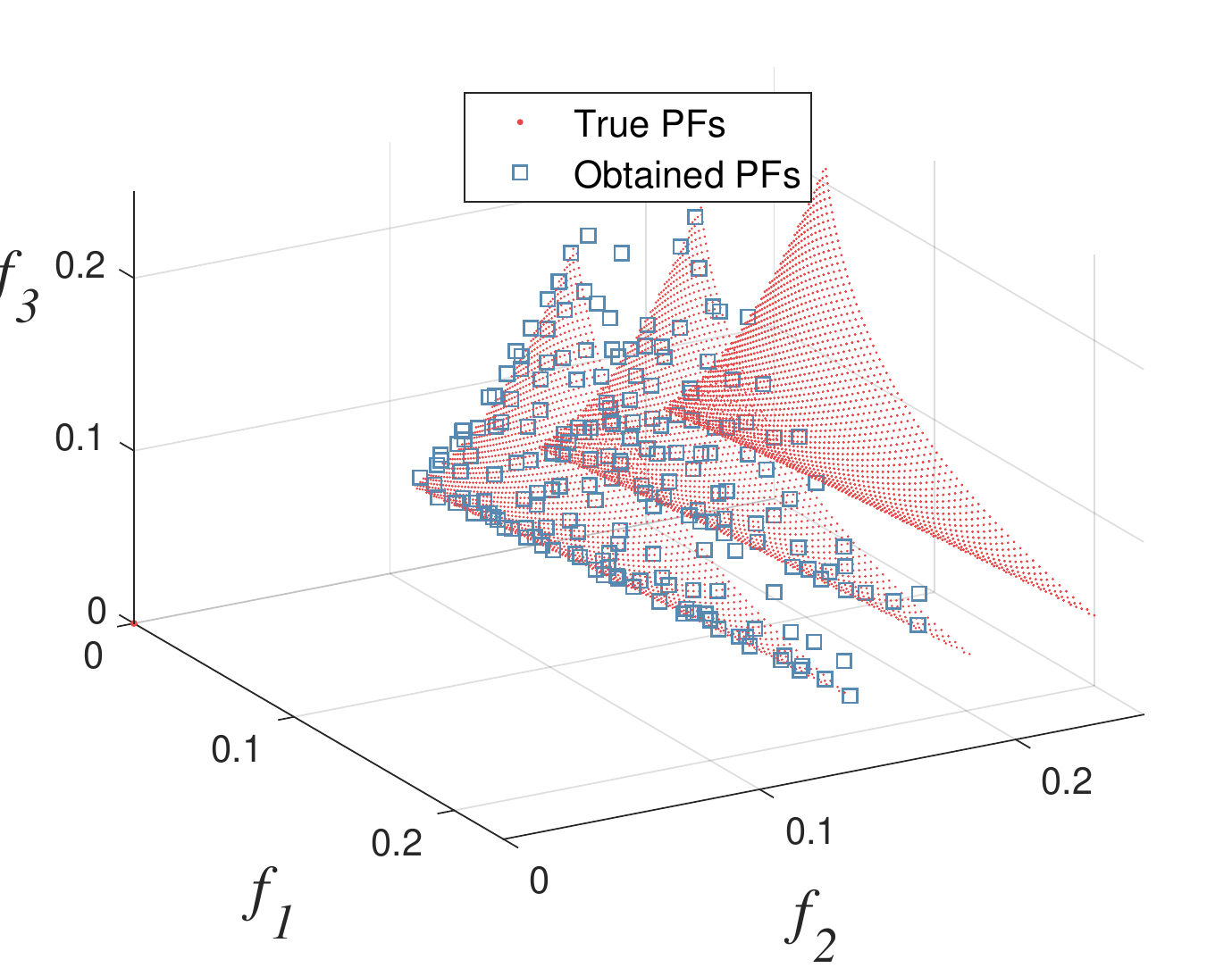}}
\subfigure[$\epsilon$=0.6]{\includegraphics[width=2.2in]{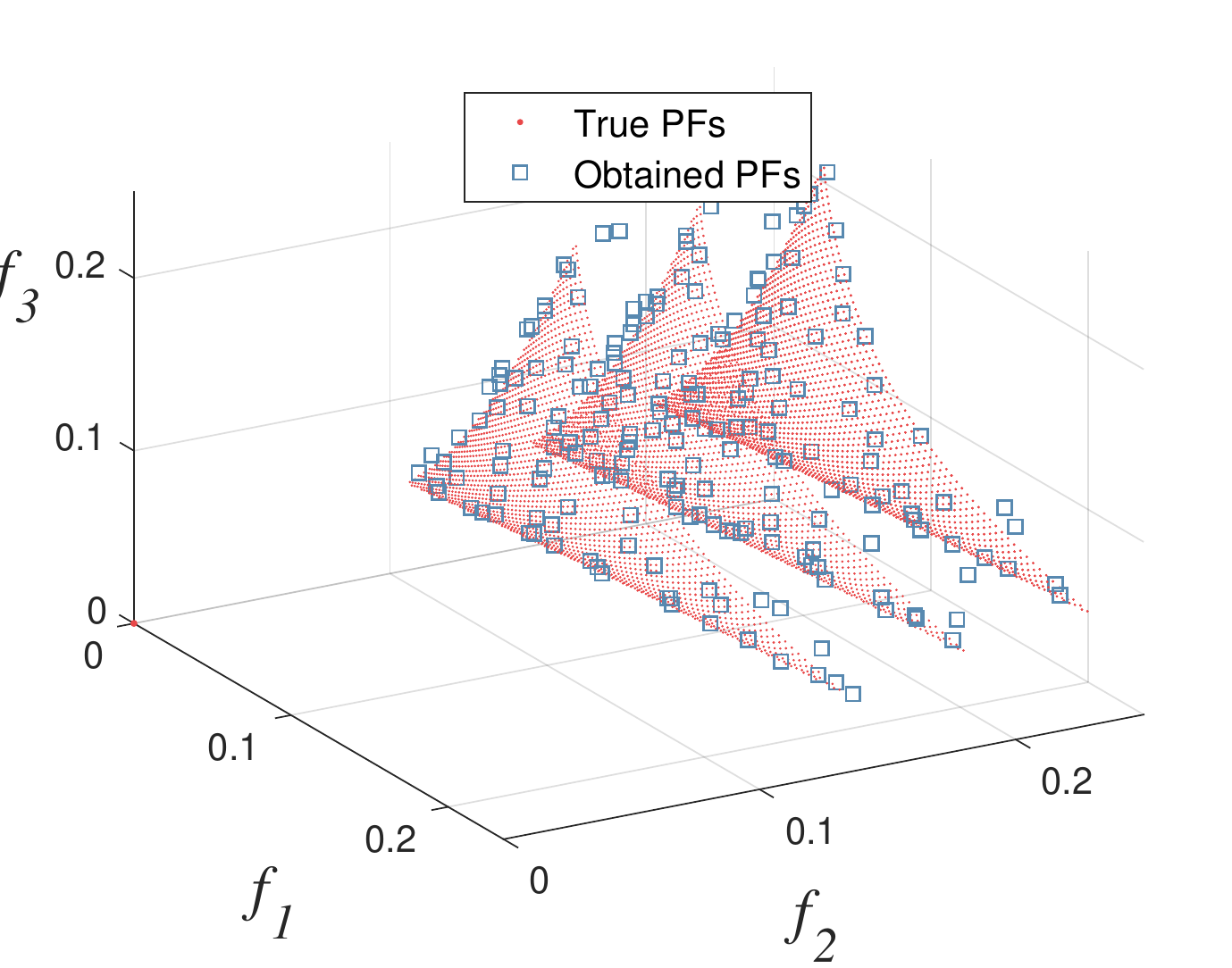}}
\caption{Distribution of the obtained solutions in the objective spaces on IDMPM3T4\_e with different $\epsilon$ values.}
\label{fig_eps}
\end{figure*}

\fref{fig_eps} shows the obtained solutions by CoMMEA with different $\epsilon$ values. From the results, we can see that $\epsilon$ can control the quality of the obtained local PFs. If we set $\epsilon=0$, then IDMPM2T4\_e only focuses on the global PF and global PSs. In this situation, CoMMEA is more likely to a normal MMEAs without considering local PSs. As $\epsilon$ increases, CoMMEA can reserve local PFs with $\epsilon$-acceptable quality. For example, when $\epsilon=0.3$, CoMMEA obtains one global PF and one local PF; when $\epsilon \geq 0.3$, CoMMEA outputs the global PF and all local PFs. Because the parameter $\epsilon$ has the same meaning as the definition of the $\epsilon$-approximate Pareto set, DMs can easily run CoMMEA according to their preferences, resulting in PSs of varying quality.

\subsection{CoMMEA's Searching behaviors}
The coevolutionary framework proposed in this study adopts two archives to appropriately balance convergence and diversity in the search process. To better analyze the searching behaviors of the two archives, we choose the multi-polygon problem with 10 decision variables and 2 objectives to present the distribution of solutions.

\begin{figure*}[tbh]
\centering
\subfigure[Stage=10\%]{\includegraphics[width=1.7in]{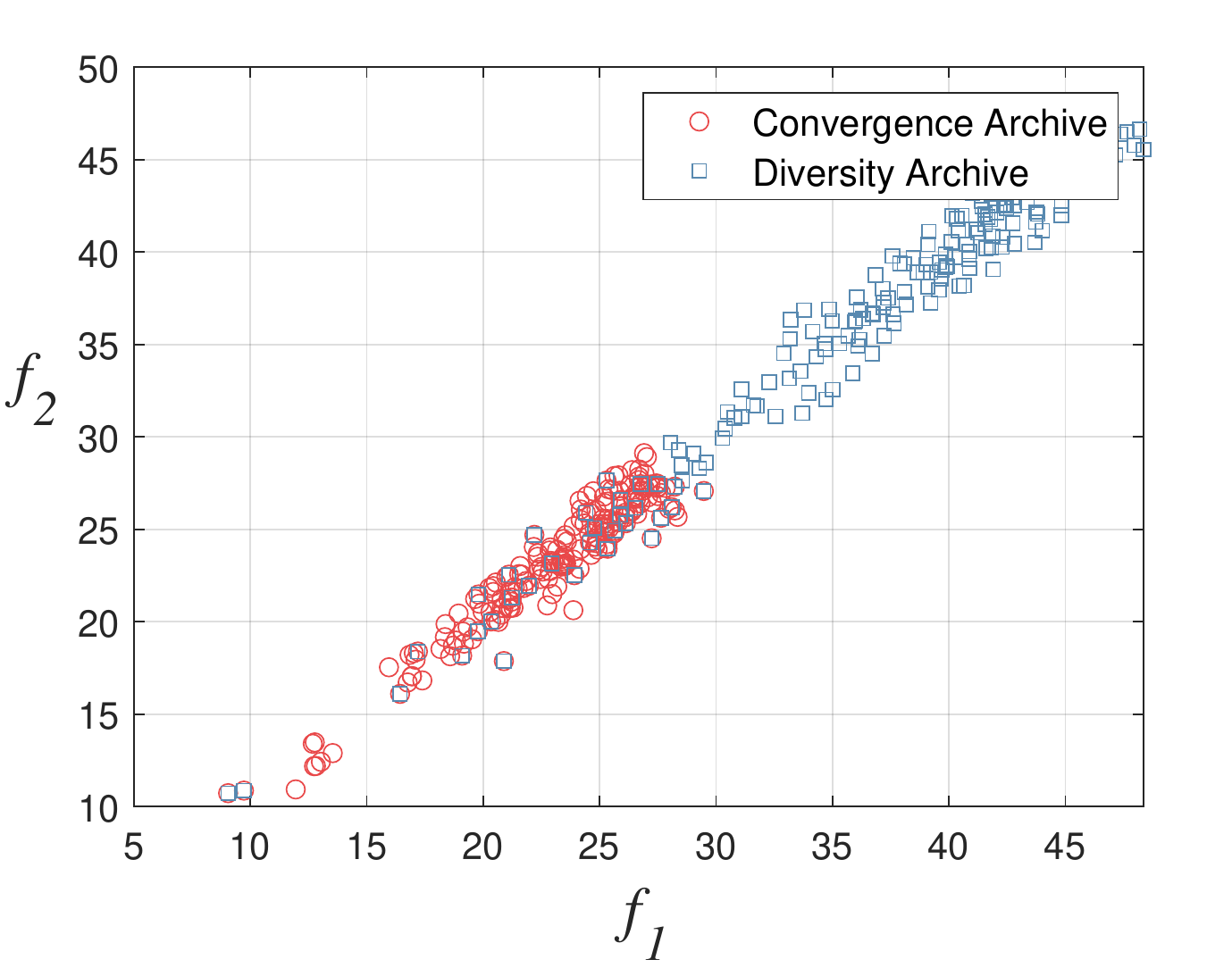}}
\subfigure[Stage=40\%]{\includegraphics[width=1.7in]{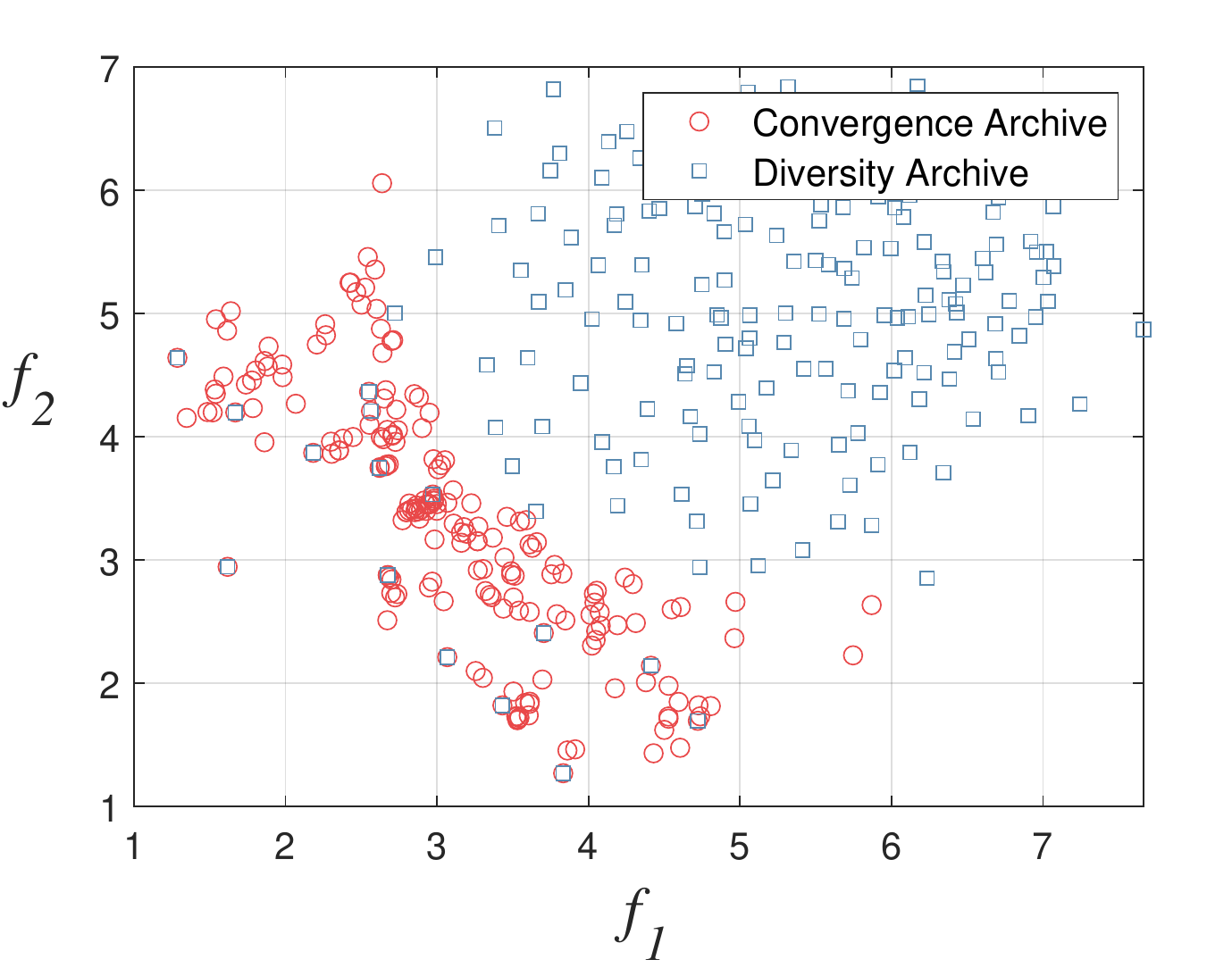}}
\subfigure[Stage=70\%]{\includegraphics[width=1.7in]{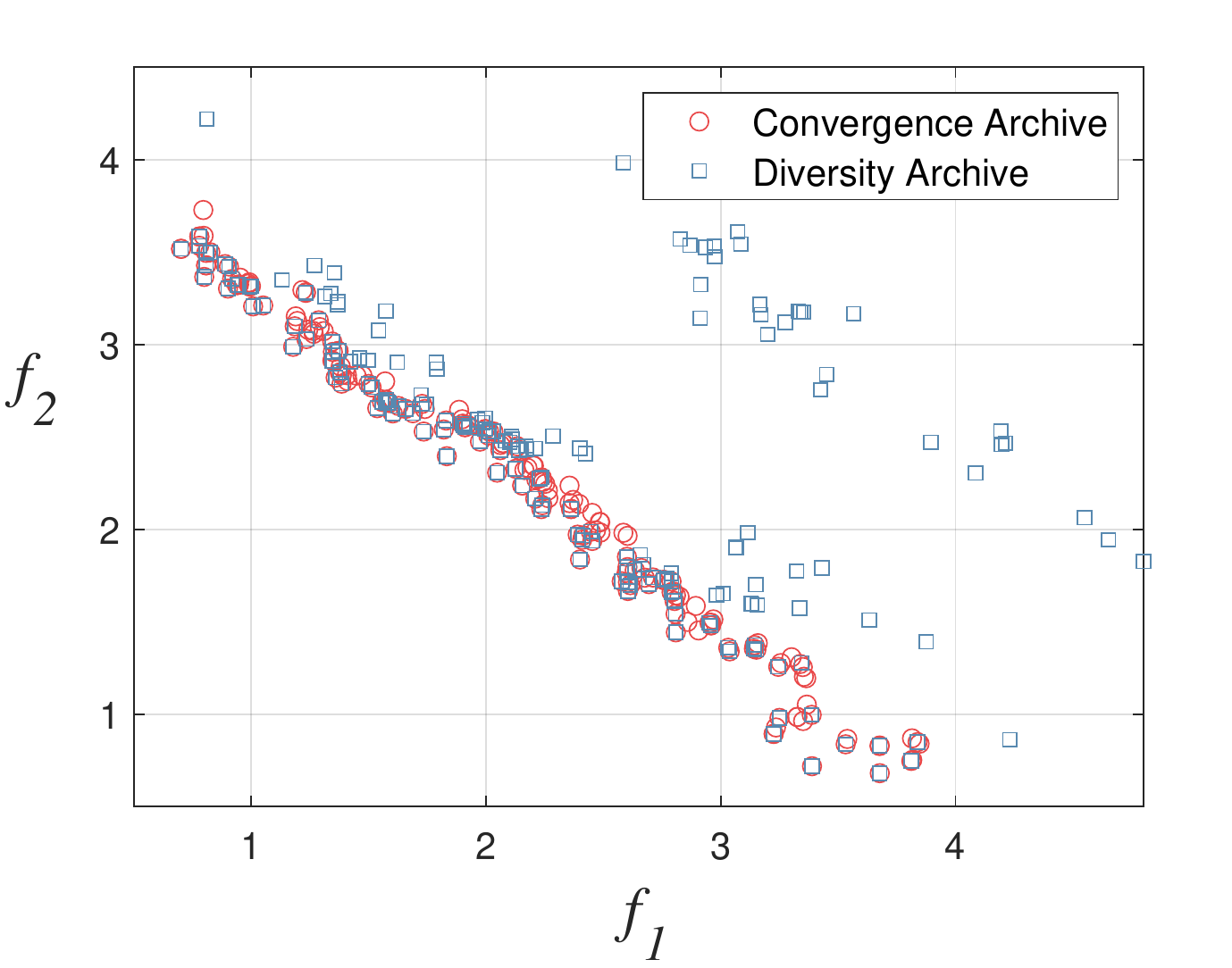}}
\subfigure[Stage=100\%]{\includegraphics[width=1.7in]{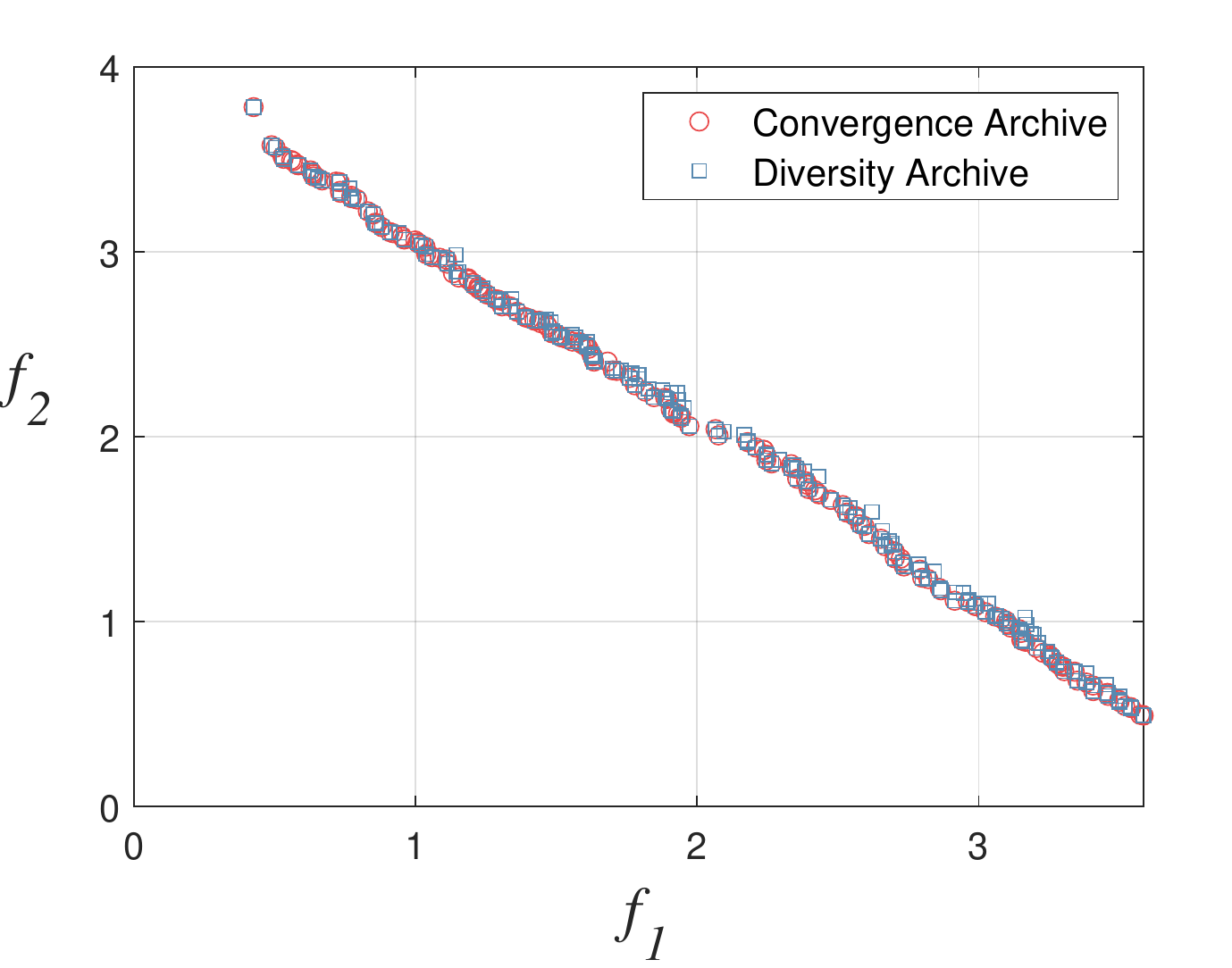}}
\subfigure[Stage=10\%]{\includegraphics[width=1.7in]{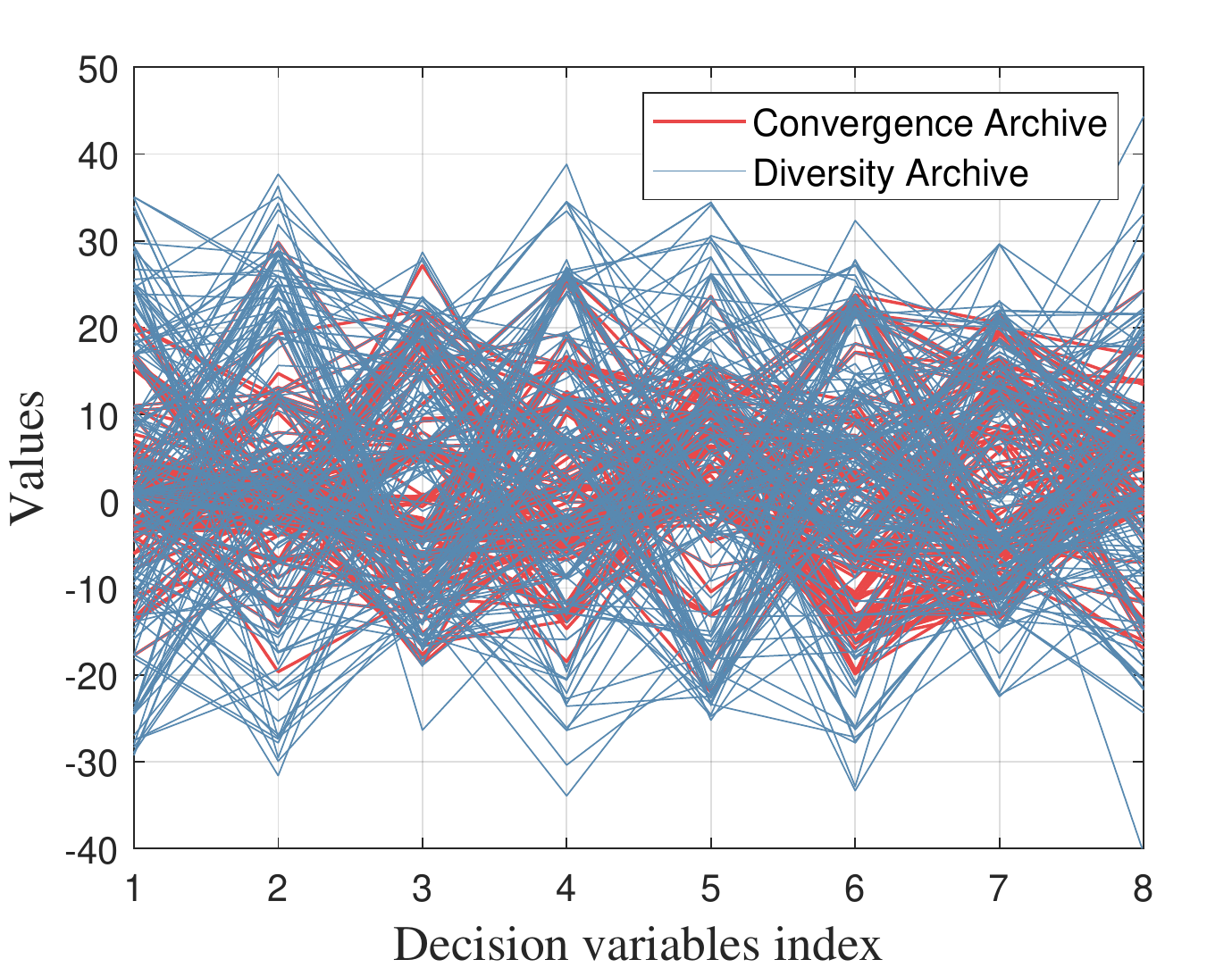}}
\subfigure[Stage=40\%]{\includegraphics[width=1.7in]{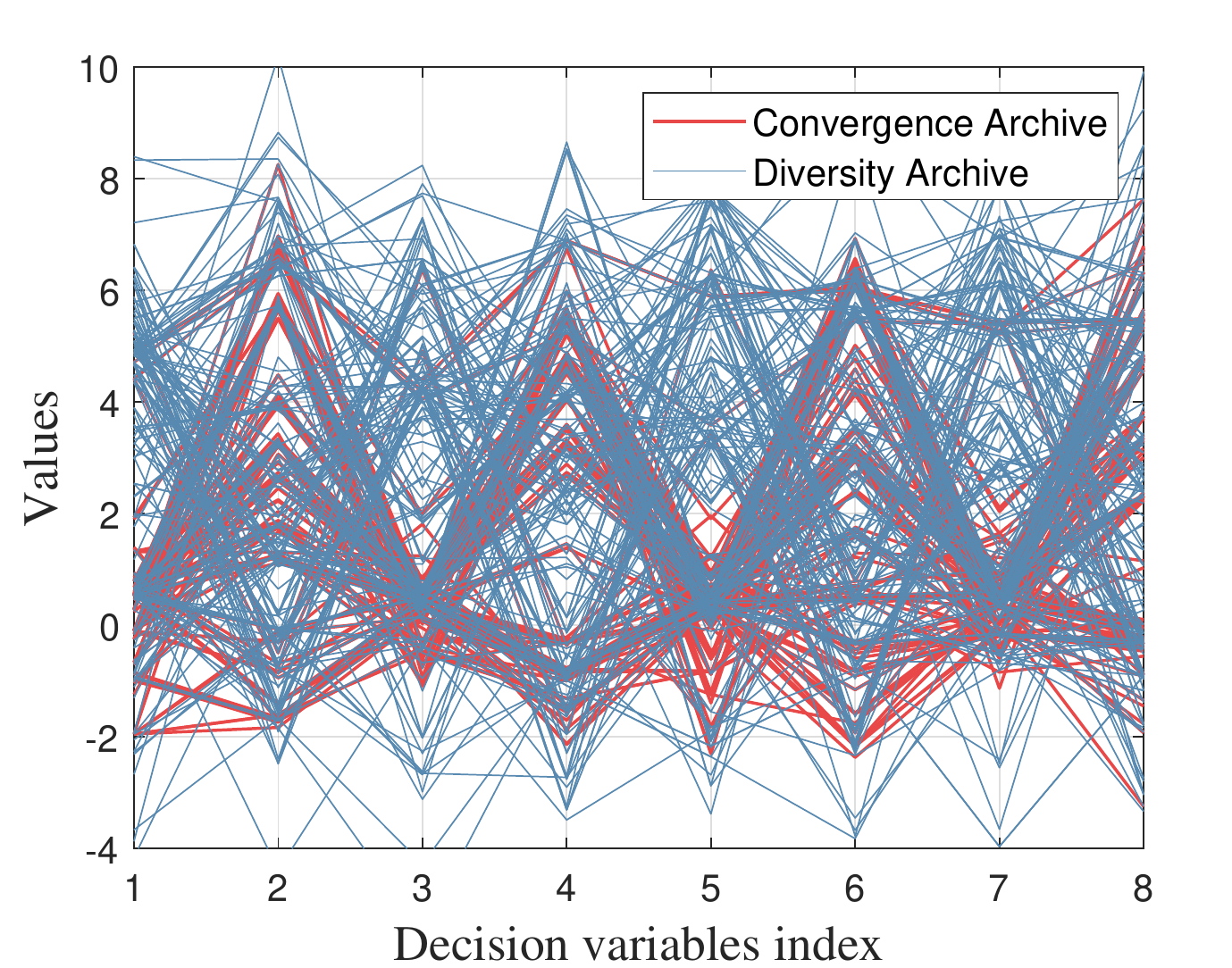}}
\subfigure[Stage=70\%]{\includegraphics[width=1.7in]{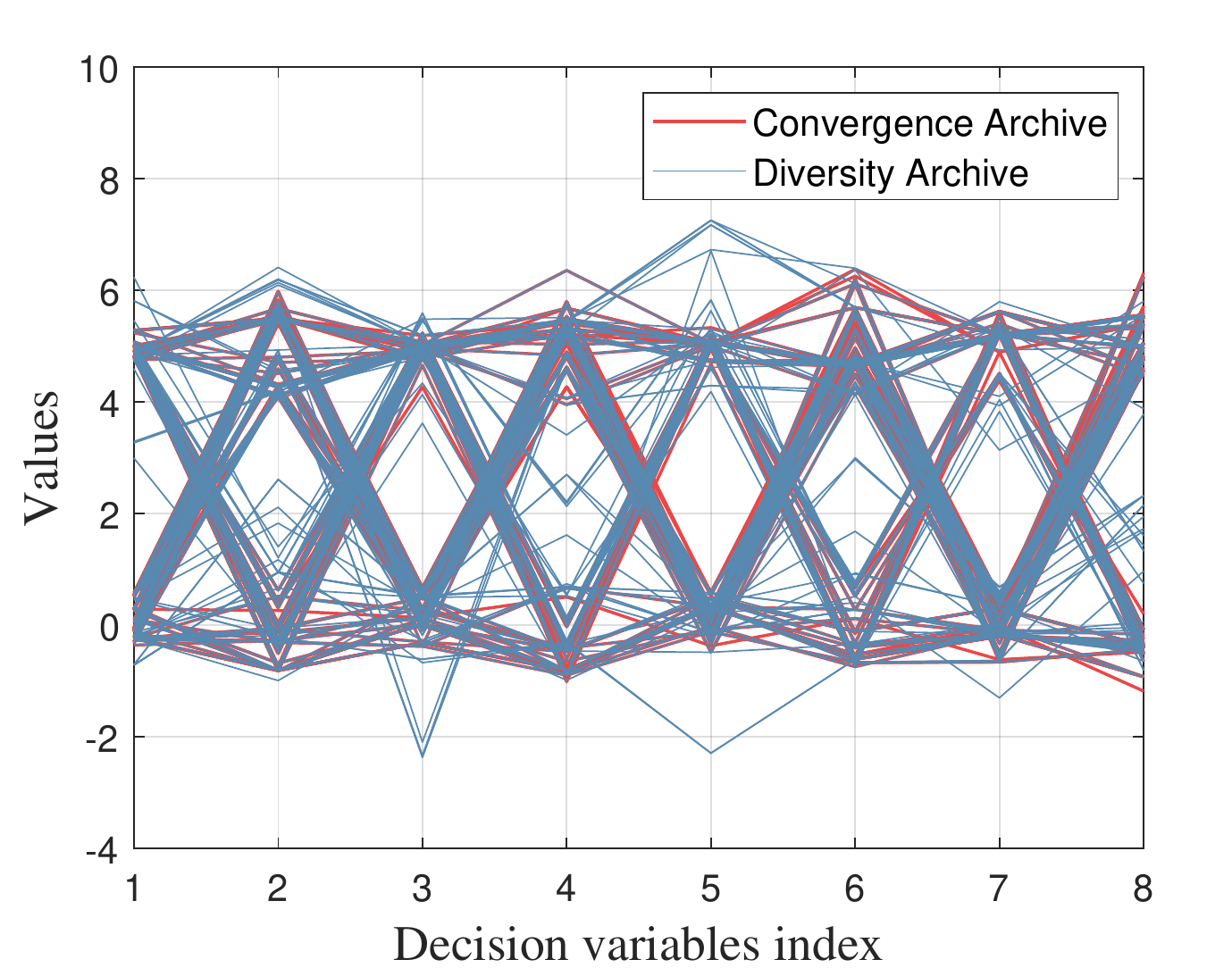}}
\subfigure[Stage=100\%]{\includegraphics[width=1.7in]{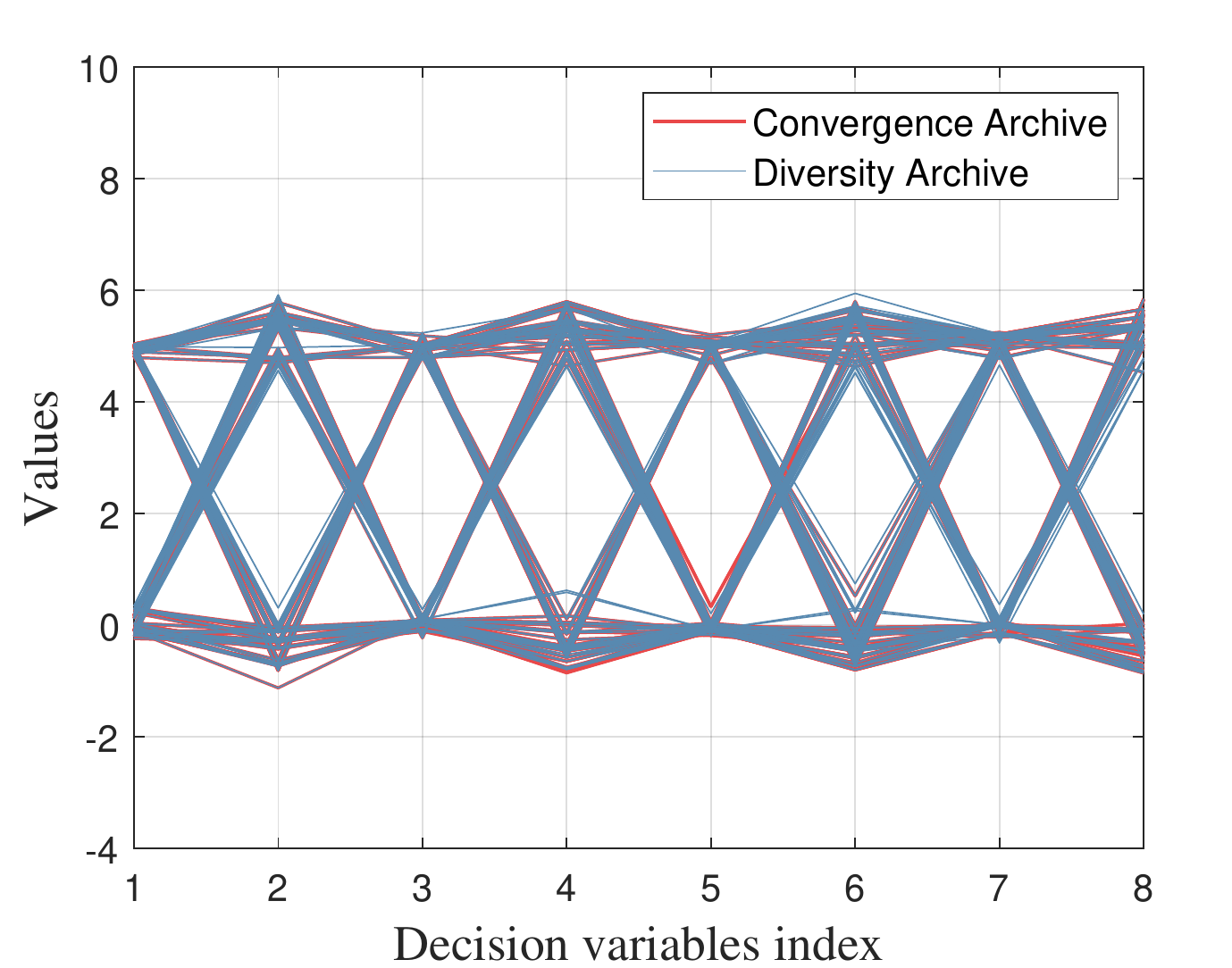}}
\caption{Distribution of solutions in the objective space (the first row) and the decision space (the last row) on multi-polygon by CoMMEA on different stages.}
\label{fig_stage}
\end{figure*}

\begin{figure}[!t]
	\centering
	\subfigure[D=4]{\includegraphics[width=1.7in]{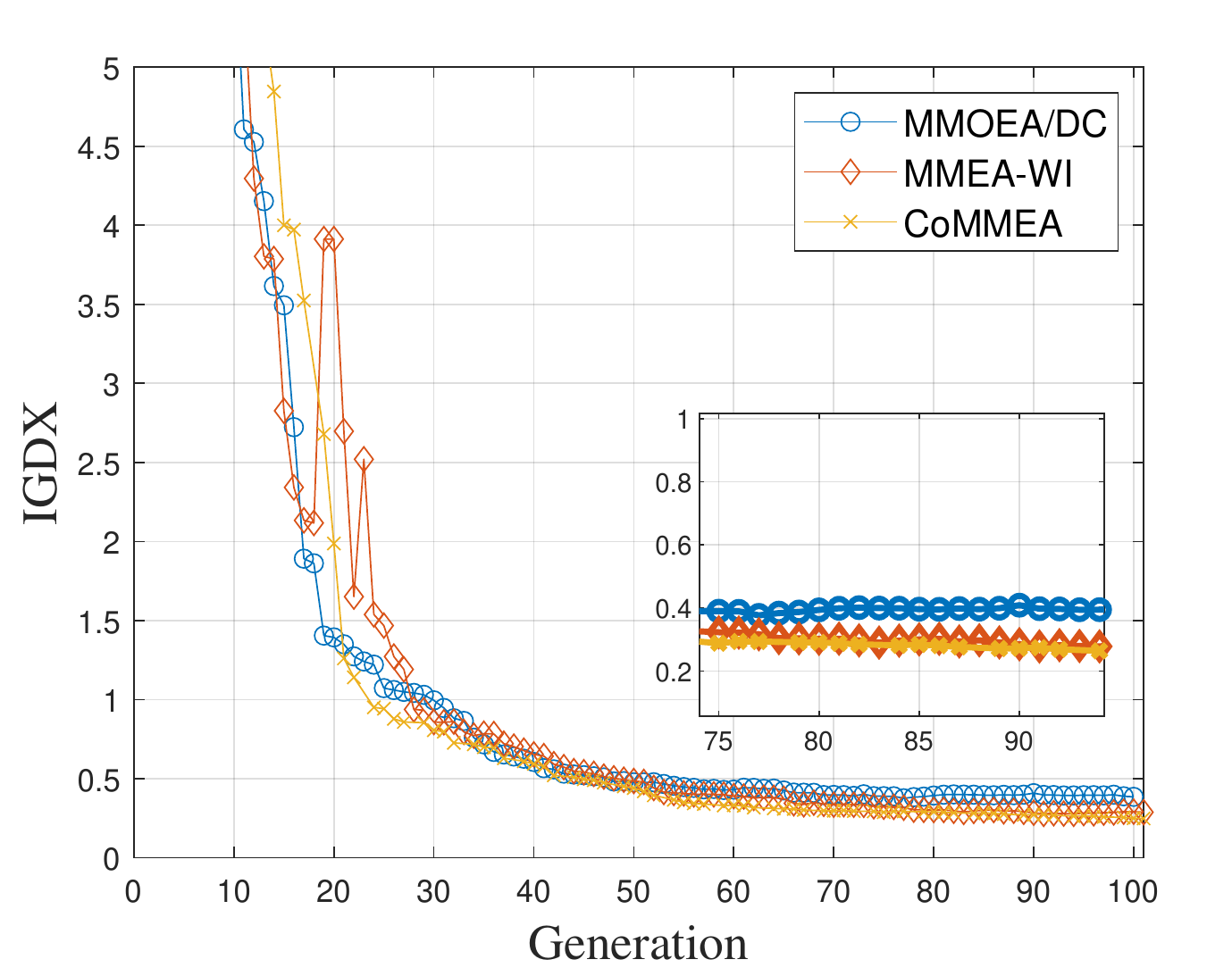}}
	\subfigure[D=8]{\includegraphics[width=1.7in]{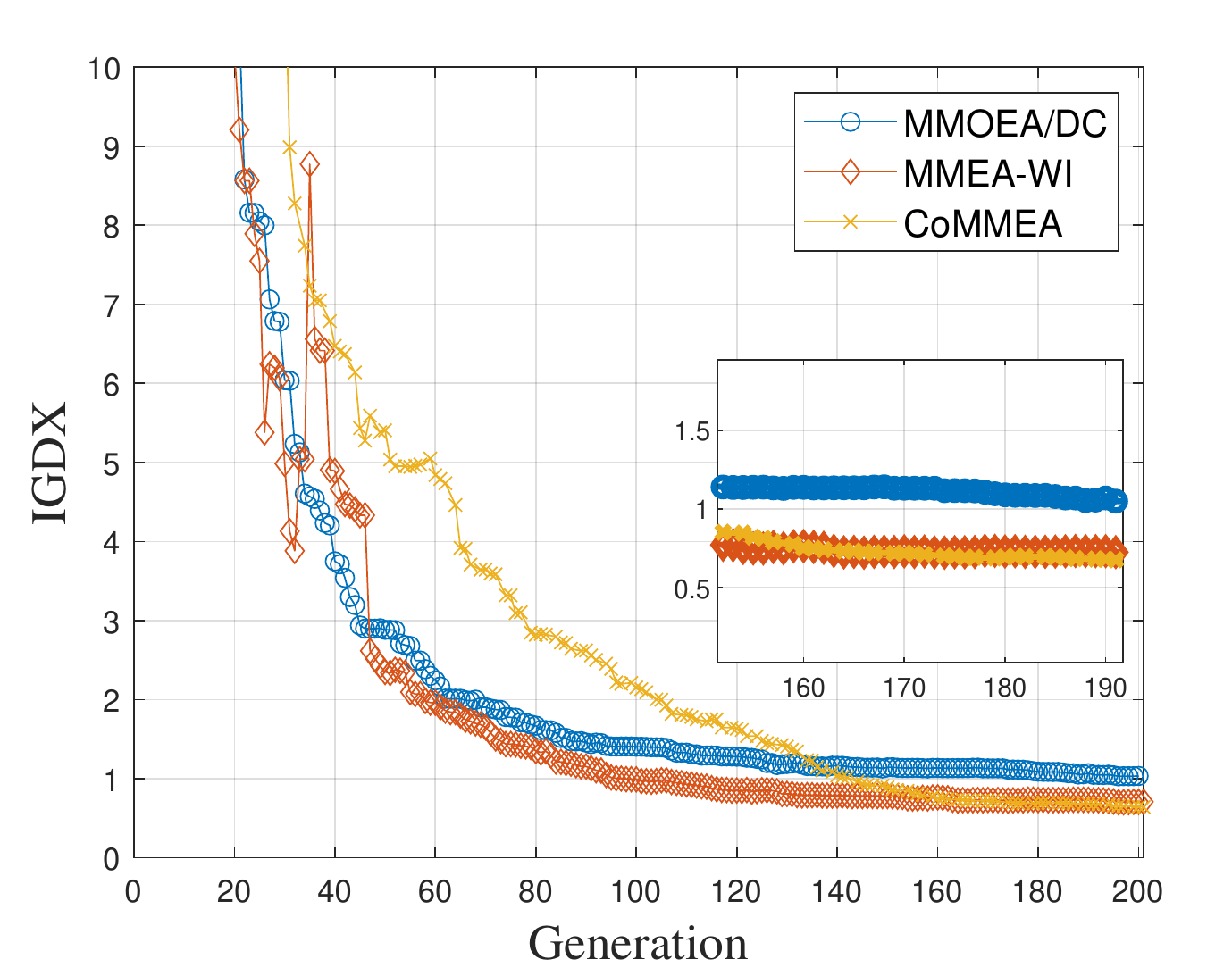}}
	\caption{Convergence curves of IGDX obtained by CoMMEA, MMEA-WI and MMOEA/DC.}
	\label{fig_conpro}
\end{figure}

\fref{fig_stage} presents the distribution of solutions in different searching stages. As we can see, in the early stage, we set $\epsilon _i$ to a large value. Therefore, solutions with poor convergence can be reserved in the diversity archive ($DA$). As a comparison, solutions in the convergence archive ($CA$) adopt the convergence-first strategy. As a result, they will quickly converge to the true PF and transfer the found PF information to the diversity archive to lead the search. 

In the middle stage, as $\epsilon _i$ decreases, the convergence quality of solutions in the diversity archive becomes better. In this stage, local PSs with $\epsilon$-acceptable quality will also be reserved. If there is no local PSs, solutions will directly converge to the true PF. On the other hand, we can find that in the convergence archive, only two global PSs (there are four global PSs) are found and reserved. In the final stage, solutions in the diversity archive bring the multiple PSs information to the convergence archive, where multiple global PSs can be obtained.

\fref{fig_conpro} presents the convergence changes in terms of IGDX obtained by the best three MMEAs on multi-polygon problems with 3 objectives. For problems with 4 and 8 decision variables, MMEA-WI and MMOEA/DC have faster convergence speeds in the early stage. This is because of the introduction of the $\epsilon$-dominance method, which is to enhance the solutions' diversity. As we can see from \fref{fig_stage} in the early stages, there are plenty of solutions distant from the known PF. As a result, the convergence speed of CoMMEA is slow in the early stages. As $\epsilon _i$ decreases, bad-quality solutions will be discarded. As a result, convergence and distribution of CoMMEA are enhanced in the latter stage.

Overall, the coevolutionary framework proposed in this study can well balance convergence and diversity. Specifically, in the former stage, the convergence archive will provide information that can lead the algorithm to converge to the true PF, while the diversity archive will enhance the population diversity to find all global and local PSs. In the later stage, both convergence and diversity archives focus on balancing the convergence and uniformity of solutions. Therefore, for GMMOPs, CoMMEA can find all equivalent PSs and obtain good convergence quality solutions for general complex MOPs.

\section{Conclusion}
\label{sec_con}

Obtaining all equivalent global PSs and high-quality local PSs is important for DMs to deal with multimodal multi-objective optimization problems. So far, the state-of-the-art MMEAs show poor convergence in dealing with problems having many decision variables, which is an obstacle for DMs to choose such algorithms. To address this issue, we proposed a coevolutionary framework that adopts two archives to coevolve simultaneously through the offspring information-sharing method. Compared to other state-of-the-art MMEAs, CoMMEA shows excellent convergence performance in dealing with complex GMMOPs, and obtains both global and local PSs according to DM's preference.

Over the last two decades, a number of studies focusing on MMOPs have been conducted. In the preliminary research on this field, to show the ability in finding multiple PSs more intuitively, almost all benchmark problems are designed with 2-3 decision variables. To achieve better results in terms of performance metrics, when designing algorithms and comparing performances, the diversity of solutions is overemphasized while the convergence ability is somehow ignored. As a result, existing MMEAs receive excellent performance in dealing with low-dimension problems but get poor results in solving complex problems, which is a kind of over-fitting. This study has made some effort in improving the convergence ability of MMEA while keeping the diversity-maintaining ability. In the future, designing more real-world-like test problems can be interesting, including large-scale problems \cite{wang2022reinforcement}, problems with difficulties in finding the true PF and problems with various constraints.

\bibliographystyle{IEEEtran}
\bibliography{newref}

\end{document}